\documentclass[10pt]{article}
\usepackage{fullpage}

\usepackage{ssArxiv}

\usepackage{microtype}

\usepackage{amsmath,amsbsy,amsgen,amscd,amssymb,amsthm,amsfonts}
\usepackage{mathtools}
\usepackage{bm}
\usepackage{enumitem}
\usepackage{comment}
\usepackage{microtype}
\usepackage{tikz}
\usepackage{url} 
\usepackage{graphicx}
\usepackage{wrapfig}
\usepackage{xspace}
\usepackage{booktabs}
\usepackage{caption}
\usepackage{subcaption}
\usepackage{braket}

\usepackage{algorithm}

\usepackage[algo2e,inoutnumbered,linesnumbered,algoruled,vlined]{algorithm2e}
\usepackage{algpseudocode}
\usepackage{bm}

\usepackage[pagebackref,breaklinks,colorlinks,citecolor=blue]{hyperref}

\usepackage[capitalize]{cleveref}

\newtheorem{remark}{Remark}
\newtheorem{proposition}{Proposition}

\newcommand{\methodname}{Q-FWAL}

\usepackage{setspace}

\usepackage[title]{appendix}

\newcommand{\norm}[1]{\|#1\|}

\newcommand{\R}{\mathbb{R}} %

\newcommand{\x}{\mathbf{x}} %
\newcommand{\X}{\mathbf{X}} %
\newcommand{\Xset}{{\cal X}} %
\newcommand{\Id}{\mathbf{I}} %
\newcommand{\W}{\mathbf{W}} %
\newcommand{\J}{\mathbf{J}} %
\newcommand{\A}{\mathbf{A}} %
\newcommand{\Pmat}{\mathbf{P}} %
\newcommand{\Q}{\mathbf{Q}} %
\newcommand{\bb}{\mathbf{b}} %
\newcommand{\s}{\mathbf{s}} %

\newcommand{\one}{\mathbf{1}} %
\newcommand{\Edge}{\mathcal{E}} %
\newcommand{\Man}{\mathcal{M}} %
\newcommand{\Pset}{\mathcal{P}} %
\renewcommand{\vec}{\mathrm{vec}} %

\newcommand{\vct}[1]{\mathbf{#1}}
\newcommand{\mtx}[1]{\mathbf{#1}}

\newcommand{\argmin}[1]{\underset{#1}{\mathrm{arg\,min}}}

\usepackage{cleveref}
\crefname{section}{\S}{\S\S}
\crefname{subsection}{\S}{\S\S}
\crefname{conj}{Conj.}{Conj.}
\crefname{assumption}{assumption}{assumptions}
\crefname{algorithm}{\text{Alg.}}{\text{Alg.}}
\crefname{assumption}{\textbf{H}}{\textbf{H}}
\crefname{equation}{\text{Eq}}{\text{Eq}}
\crefname{definition}{\text{Dfn.}}{\text{Dfn.}}
\crefname{lemma}{\text{Lemma}}{\text{Lemma}}
\crefname{dfn}{\text{Dfn.}}{\text{Dfn.}}
\crefname{thm}{\text{Thm.}}{\text{Thm.}}
\crefname{tab}{\text{Tab.}}{\text{Tab.}}
\crefname{fig}{\text{Fig.}}{\text{Fig.}}
\crefname{table}{\text{Tab.}}{\text{Tab.}}
\crefname{figure}{\text{Fig.}}{\text{Fig.}}

\newcommand{\eg}{\textit{e}.\textit{g}. }
\newcommand{\etal}{\textit{et al}. }
\newcommand{\ie}{\textit{i}.\textit{e}. }
\newcommand{\cf}{\textit{cf}.} 

\renewcommand{\paragraph}[1]{{\vspace{1mm}\noindent \bf #1}.}

\newenvironment{brsm}{%
  \bigl[ \begin{smallmatrix} }{%
  \end{smallmatrix} \bigr]}

\begin{document}

\title{Q-FW: A Hybrid Classical-Quantum Frank-Wolfe for Quadratic Binary Optimization}

 \author{\name Alp Yurtsever \email{alp.yurtsever@umu.se} \\
      \addr{Umeå University, Sweden} \\[0.5em]
   \name Tolga Birdal \email{t.birdal@imperial.ac.uk} \\
      \addr{Imperial College London, UK} \\[0.5em]
   \name Vladislav Golyanik \email{golyanik@mpi-inf.mpg.de} \\
      \addr{MPI for Informatics, Germany}
 }

\maketitle

\begin{abstract}
    We present a hybrid classical-quantum framework based on the Frank-Wolfe algorithm, Q-FW, for solving quadratic, linearly-constrained, binary optimization problems on quantum annealers (QA). The computational premise of quantum computers has cultivated the re-design of various existing vision problems into quantum-friendly forms. Experimental QA realisations can solve a particular non-convex problem known as the quadratic unconstrained binary optimization (QUBO). Yet a naive-QUBO cannot take into account the restrictions on the parameters. To introduce additional structure in the parameter space, researchers have crafted ad-hoc solutions incorporating (linear) constraints in the form of regularizers. However, this comes at the expense of a hyper-parameter, balancing the impact of regularization. To date, a true constrained solver of quadratic binary optimization (QBO) problems has lacked. Q-FW first reformulates constrained-QBO as a copositive program (CP), then employs Frank-Wolfe iterations to solve CP while satisfying linear (in)equality constraints. This procedure unrolls the original constrained-QBO into a set of unconstrained QUBOs all of which are solved, in a sequel, on a QA. We use D-Wave Advantage QA to conduct synthetic and real experiments on two important computer vision problems, graph matching and permutation synchronization, which demonstrate that our approach is effective in alleviating the need for an explicit regularization coefficient. 
\end{abstract}

\section{Introduction}\label{sec:intro}
Combinatorial optimization is at the heart of computer vision (CV). In a variety of applications such as structure-from-motion (SfM)~\cite{schonberger2016structure}, SLAM~\cite{murORB2}, 3D reconstruction~\cite{dai2017bundlefusion}, camera re-localization~\cite{Sattler12imageretrieval}, image retrieval~\cite{li2015pairwise} and 3D scan stitching~\cite{huber2003fully,deng2018ppf}, correspondences serve as a powerful proxy to visual perception. In many problems, correspondences are defined over two or multiple point sets and can be encoded as \emph{permutation matrices} that are binary assignment operators. Recovering permutations from observations involve solving NP-hard combinatorial problems. As a remedy, scholars have opted to \emph{relax} those problems to arrive at tractable albeit suboptimal solutions~\cite{xiang2020efficient,birdal2019probabilistic,caetano2009learning,kezurer2015tight}. However, recent advances in computer hardware urges us to re-visit our approaches.

Quantum computers (QCs) harness the collective properties of quantum states, such as superposition, interference and entanglement to perform calculations \cite{nielsen}. Thanks to the use of a more advanced physics, QCs can offer theoretical improvements in the face of complexity classes that are challenging to handle today \cite{Shor1994}.
With the experimental realization of quantum supremacy~\cite{arute2019quantum}, we are now more confident that practical quantum computing is right around the corner. 

A particular quantum computational model, 
known as 
\emph{Adiabatic Quantum Computing} (AQC), is based on the  adiabatic theorem of quantum mechanics  \cite{BornFock1928}. 
Closely related to it is \emph{Quantum Annealing} (QA), 
which is a quantum optimization method (AQC-type) 
that implements a qubit-based quantum system described by the Ising model~\cite{KadowakiNishimori1998}. 
Albeit restricted, experimental realisations of QA, such as DWave~\cite{Dattani2019}, 
can solve non-convex, \emph{quadratic unconstrained binary optimization} (QUBO) problems, without resorting to continuous relaxations. 
This premise of AQC and QA has led to the emergence of \emph{quantum computer vision} (QCV), 
where researchers started to port existing computer vision problems into forms amenable to quantum computation~\cite{hu2019quantum,LiGhosh2020,pointCorrespondence,SeelbachBenkner2020,zaech2022adiabatic,birdal2021quantum}.

Even though employing QA to solve CV problems has shown benefit\footnote{Quantum computers are still in early stages. However, a diverse set of CV experiments present optimistic predictions regarding the future.}, a large body of computer vision algorithms rely on some form of (in)equality constraints to be incorporated. For example, estimating correspondences require solving QBOs for permutations and not for arbitrary binary vectors.
To this end, the state of the art QCV methods either use a regularization with cherry picked coefficients~\cite{birdal2021quantum} or resort to heuristics for auto-controlling the impact of the constraints~\cite{SeelbachBenkner2020,zaech2022adiabatic}. Unfortunately, none of these approaches are optimal and jeopardize the solution quality guarantees of quantum computers.

In this paper, we address the above issue of incorporating (in)equality constraints and introduce Quantum-Frank Wolfe (Q-FW), a Frank-Wolfe framework for satisfying linear (in)equality constraints in a QBO problem. Q-FW is based on an equivalent \emph{copositive programming} formulation of constrained QBO and involves iteratively solving a sequence of classical, unconstrained QUBOs. 
At its core, on a classical computer, Q-FW employs one of the two variants of FW tailored for solving CP problems; FW with augmented Lagrangian (FWAL) \cite{yurtsever2019conditional} or FW with quadratic penalty (FWQP) \cite{yurtsever2018conditional}. At every iteration, these methods identify an update direction by minimizing a linear approximation of a penalized proxy of the objective function. Q-FW formulates this linear minimization as a QUBO and 
obtains the update direction via QA.  %
We then take a small step in this update direction. In addition, FWAL maintains a dual variable, updated by a small gradient step for improved numerical performance. 

Thanks to the convexity inherent in CP, Q-FW converges to the global minimum regardless the choice of its algorithm parameters.  
By virtue of the exact copositive-reformulation, our solutions are oftentimes near the true global minimum, obtained via an exhaustive search in small problems. We deploy Q-FW on multiple computer vision tasks of permutation synchronization and graph matching, which both have wide applicability. Our contributions are:
\begin{itemize}[noitemsep,leftmargin=\parindent,topsep=0.1em]
    \item We introduce Q-FW, an adaptation of the classical FW algorithm for solving copositive programs on a hybrid classical-quantum computing system. %
    \item We solve the challenging QUBO sub-problems using an actual experimental realisation of a quantum annealer (QA), DWave Advantage  4.1~\cite{DWAVE,Dattani2019}.
    \item We tackle both graph matching and permutation synchronization problems and obtain excellent results on both synthetic and real benchmarks.
\end{itemize}
Our evaluations confirm the theoretical advantages of Q-FW: Q-FW is robust, can solve larger problems than brute-force search, can exactly satisfy (in)equality constraints and enjoys a tight copositive relaxation. We plan to release our software package upon publication.

\vspace{-3mm}\section{Related Work}\label{sec:relwork}
\vspace{-3mm}
Our approach relates to different methods both in classical optimization and quantum computer vision. In this section, we review the most related works in QCV, copositive programming and FW. 

\paragraph{Quantum computer vision (QCV)} QCV encompasses hybrid classical-quantum methods with parts solved on a gate-based quantum computer or a quantum annealer. This young field seeks to identify how challenging problems can be formulated for and benefit from quantum hardware. While it remained predominantly theoretical at early stages \cite{Neven2008arXivRecognition,chin2020quantum}, QCV methods from various domains were evaluated on real quantum hardware during the recent few years, including image classification \cite{Neven2012,NguyenKenyon2019,Cavallaro2020}, object detection \cite{LiGhosh2020}, graph matching \cite{SeelbachBenkner2020}, mesh alignment \cite{benkner2021q}, robust fitting \cite{doan2022hybrid} and permutation synchronisation \cite{birdal2021quantum}. 

Some of the proposed algorithms require additional constraints formulated as weighted linear terms (Lagrange multipliers) \cite{SeelbachBenkner2020,birdal2021quantum,zaech2022adiabatic}. Such conditions rectify the original unconstrained objective and preserve the QUBO form consumable by modern QA. However, since the linear constraints modify the problem's energy landscape, the corresponding weights have to be chosen with care; too high or too low weights can significantly decrease the probability of measuring optimal solutions after the sampling. Birdal~\etal~\cite{birdal2021quantum} select the weights with a time-consuming grid search (for small problem instances). Benkner \etal~\cite{SeelbachBenkner2020} derive lower bounds on the rectification weights for the quadratic assignment problem. Both policies have a common limitation: The determined weights are problem-specific and do not generalise to other problems. Moreover, even problems of the same type and size can demand new multipliers. 

In contrast to existing methods, our unified policy does not require selecting the weights of linear terms in advance. 
Similar to Q-Match \cite{benkner2021q}, our method is iterative: a sequence of optimisation tasks are solved on QPU; in each iteration, the control is returned to CPU to define a follow-up QUBO until convergence. 
Q-Match~\cite{benkner2021q} update its solutions via a series of permutation-ness-preserving directions (collections of 2-cycles). 
Its policy does not generalise to other problems, arbitrary solution encodings and weighted linear constraints, as our method does. 

\paragraph{Copositive programming (CP)} 
CP is a subfield of convex optimization concerned with optimizing a linear objective under affine constraints over the cone of copositive matrices, or its dual cone, the cone of completely positive matrices. By definition, a matrix $\mtx{X} \in \mathbb{R}^{n \times n}$ is said to be copositive if its quadratic form is nonnegative on the first orthant (\ie\!, $\vct{z}^\top \mtx{X} \vct{z} \geq 0$ for all $\vct{z} \in \mathbb{R}^n_+$) and completely positive if $\mtx{X} \in \mathrm{conv} \{\vct{x}\vct{x}^\top: \vct{x} \in \mathrm{R}^n_+\}$. 
Compared to semidefinite programming, CP provides a tighter relaxation of quadratic problems \cite{quist1998copositive}. 
However, despite its convexity, solving a CP problem is NP-Hard \cite{bomze2000copositive}. Several NP-Hard problems in quadratic and combinatorial optimization are subsets of CP, including the binary quadratic problems \cite{burer2009copositive}, problems of finding stability and chromatic numbers of a graph \cite{de2002approximation,dukanovic2010copositive}, quadratic assignment problem \cite{povh2009copositive}, and training of vector-output RELU networks~\cite{sahiner2020vector}. We refer to the excellent surveys \cite{dur2010copositive,dur2021conic} and references therein for more details. %

\paragraph{Frank Wolfe (FW)}
FW (also known as conditional gradient method or CGM) is a classical method in convex optimization dating back to 1956 \cite{frank1956algorithm}. Initially, the method is proposed for minimizing a convex quadratic loss function over a polytope. The analysis is extended in \cite{Levitin1966} to minimize a generic smooth and convex objective over an arbitrary convex and compact set. The eccentric feature of FW is that it does not require a projection step, which is in stark contrast with most other methods for constrained optimization, and it makes FW efficacious for problems where projection is computationally prohibitive. 
FW is demonstrated as an effective method for optimization over simplex \cite{Clarkson2010} or spactrahedron domains \cite{hazan2008sparse}. %
We refer to \cite{jaggi2013revisiting} for convergence analysis of FW and a detailed discussion on its applications, and to \cite{Bomze2021frankwolfe} for a review on recent advances in FW. 

The original form of FW is not suitable to tackle affine equality constraints present in our CP formulation. Instead, we consider two design variants of FW: %
FWQP \cite{yurtsever2018conditional}, which  equips FW with a quadratic penalty strategy for affine constraints; and FWAL \cite{yurtsever2019conditional}, which extends FWQP for an augmented Lagrangian penalty. Our choice is inspired by \cite{yurtsever2021scalable} using FWAL for solving semidefinite programs. We adopt a similar approach for solving CPs.

In what follows, we first formulate QBO as an instance of the more general copositive program class \S\ref{sec:problem}. We then provide our Q-FW framework for solving copositive programs in a generic way (\S\ref{sec:QFW}). Finally, we cast graph matching (\S\ref{sec:QGM}) and permutation synchronization (\S\ref{sec:QPS}) tasks as instances of QBOs with equality constraints, which Q-FW could solve effectively. %
\section{Problem Formulation}\label{sec:problem}

This section presents our model problem, a quadratic binary optimization (QBO) with affine (in)equality constraints\footnote{Throughout the paper we concentrate on the equality constraints and provide a simple modification to satisfy inequality constraints in our supplementary material.}, and an equivalent copositive program outlined in \cite{burer2009copositive}. 

We assume that the problems are presented in the following form:
\begin{equation}\label{eqn:qubo}
\min_{\vct{x} \in \mathbb{Z}_2^n} ~~ \vct{x}^\top \mtx{Q} \vct{x} + 2\,\vct{s}^\top \vct{x} \quad \text{subject to} \quad \vct{a}_i^\top \vct{x} = b_i, ~~~ i = 1,2,\ldots,m,
\end{equation}
where $\vct{x} \in \mathbb{Z}_2^n$ is the binary valued decision variable, 
$\mtx{Q} \in \mathbb{R}^{n\times n}$ and $\vct{s} \in \mathbb{R}^{n}$ are the quadratic and linear cost coefficients, and $\{(\vct{a}_i,b_i) \in \mathbb{R}^n \times \mathbb{R}\}$ are the constraint coefficients. We assume $b_i \geq 0$ without loss of generality. Throughout, we treat $\mtx{Q}$ as a symmetric matrix since \eqref{eqn:qubo} is invariant under symmetrization of $\mtx{Q}$:
\begin{equation}
    \vct{x}^\top \mtx{Q} \vct{x} = \tfrac{1}{2} \vct{x}^\top \mtx{Q} \vct{x} + \tfrac{1}{2} (\vct{x}^\top \mtx{Q} \vct{x})^\top = \vct{x}^\top ( \tfrac{1}{2}\mtx{Q} +  \tfrac{1}{2}\mtx{Q}^\top) \vct{x}.
\end{equation}
One can also drop the linear term $\vct{s}^\top \vct{x}$ from the objective, because we can translate it into the quadratic term: Given that $\vct{x}$ is binary valued, $\vct{s}^\top \vct{x} = \vct{x}^\top \mathrm{Diag}(\vct{s}) \, \vct{x}$. 

To reformulate this problem, consider the rank-one completely positive matrix $\mtx{X} = \vct{x}\vct{x}^\top \in \mathbb{Z}_2^{n\times n}$. Since $\vct{x}$ is binary valued, we have $\mathrm{diag}(\mtx{X}) = \vct{x}$. 
Then, the quadratic objective in \eqref{eqn:qubo} can be cast as a linear function of $\mtx{X}$:
\begin{equation}
    \vct{x}^\top \mtx{Q} \vct{x} = \mathrm{Tr}(\vct{x}^\top \mtx{Q} \vct{x}) = \mathrm{Tr}(\mtx{Q} \, \vct{x}\vct{x}^\top ) = \mathrm{Tr}(\mtx{Q} \mtx{X}). 
\end{equation}
Similarly, we rewrite affine constraints from problem~\eqref{eqn:qubo} by using 
\begin{equation}
\begin{aligned}
    \vct{a}_i^\top \vct{x} = b_i 
    & \iff (\vct{a}_i^\top \vct{x})^2 = b_i^2 \\
    & \iff \mathrm{Tr}(\mtx{A}_i\mtx{X}) = b_i^2 , ~~ \text{where} ~~ \mtx{A}_i := \vct{a}_i \vct{a}_i^\top,
\end{aligned}
\end{equation}
which holds true since $(\vct{a}_i^\top \vct{x})^2 = \vct{x}^\top \vct{a}_i \vct{a}_i^\top \vct{x} = \mathrm{Tr}(\vct{x}^\top \vct{a}_i \vct{a}_i^\top \vct{x}) = \mathrm{Tr}(\vct{a}_i \vct{a}_i^\top \vct{x} \vct{x}^\top)$. 

Now, we reformulate problem \eqref{eqn:qubo} as follows:
\begin{equation}
\begin{aligned}
\min_{\vct{x}, \mtx{X}} ~~ \mathrm{Tr}(\mtx{Q}\mtx{X}) + 2\,\vct{s}^\top \vct{x} \quad \text{subject to} \quad 
& \vct{a}_i^\top \vct{x} = b_i, ~~~ i = 1,2,\ldots,m, \\[-0.25em]
& \mathrm{Tr}(\mtx{A}_i \mtx{X}) = b_i^2, ~~~ i = 1,2,\ldots,m, \\[0.25em]
& \mtx{X} = \vct{x}\vct{x}^\top, ~~ \text{and} ~~ \vct{x}\in \mathbb{Z}_2^n.
\end{aligned}
\end{equation}
By replacing the nonconvex nonlinear constraint $\{\mtx{X} = \vct{x}\vct{x}^\top, \vct{x}\in \mathbb{Z}_2^n\}$ with  
\begin{equation}
    \mathrm{diag}(\mtx{X}) = \vct{x}, \text{~~and~~}
    \begin{bmatrix}
    1 & ~ \vct{x}^\top \\
    \vct{x} & ~ \mtx{X}~
    \end{bmatrix}
    \in \Delta^{n+1} ~ \text{where} ~ \Delta^n := \mathrm{conv}\{\vct{x}\vct{x}^\top: \vct{x} \in \mathbb{Z}_2^n\},
\end{equation}
we get a CP problem:
\begin{equation}\label{eqn:copo}
\begin{aligned}
\min_{\vct{x}, \mtx{X}} ~~ \mathrm{Tr}(\mtx{Q}\mtx{X}) \quad \text{subject to} \quad 
& \vct{a}_i^\top \vct{x} = b_i, ~~~ i = 1,2,\ldots,m, \\[-0.25em]
& \mathrm{Tr}(\mtx{A}_i \mtx{X}) = b_i^2, ~~~ i = 1,2,\ldots,m, \\
& \mathrm{diag}(\mtx{X}) = \vct{x}, \text{~and~} \begin{bmatrix}
    1 & ~ \vct{x}^\top \\
    \vct{x} & ~ \mtx{X}~
    \end{bmatrix}
    \in \Delta^{n+1}.
\end{aligned}
\end{equation}
This reformulation is tight, see Theorem~2.6 in \cite{burer2009copositive} for the technical derivation. Our numerical experiments demonstrate the tightness of this reformulation empirically for the graph matching and permutation synchronization problems. %

\paragraph{Compact notation}
We introduce a compact notation for problem \eqref{eqn:copo} for convenience. Let $p = n+1$, denote the new decision variable by $\mtx{W} \in \Delta^p$, and introduce a new cost matrix $\mtx{C} = \begin{brsm}
    0 & ~ \vct{s}^\top \\
    \vct{s} & ~ \mtx{Q}~
\end{brsm}
$.
Further, let $d = 2m + n +1$ and introduce a linear map $\mathcal{A}:\mathbb{R}^{p \times p} \to \mathbb{R}^d$ and vector $\vct{v}\in\mathbb{R}^d$ combining all affine constraints in problem \eqref{eqn:copo}, including $\{\vct{a}_i^\top \vct{x} = b_i\}$, $\{\mathrm{Tr}(\mtx{A}_i \mtx{X}) = b_i^2\}$, $\mathrm{diag}(\mtx{X}) = \vct{x}$, and $W_{1,1} = 1$.

In this notation, problem \eqref{eqn:copo} becomes
\begin{equation}\label{eqn:copo-compact}
\begin{aligned}
\min_{\mtx{W} \in \Delta^p} ~ \mathrm{Tr}(\mtx{C}\mtx{W}) \quad \text{subject to} \quad 
& \mathcal{A}\mtx{W} = \vct{v}.
\end{aligned}
\end{equation}
This is a convex optimization problem, but it is NP-Hard because of the complete positivity constraint. %

\section{Quantum Frank-Wolfe  (Q-FW)}\label{sec:QFW}
In the light of the copositive reformulation above, we now develop the main algorithm for solving a constrained-QBO. 
We describe the algorithm with FWAL. FWQP is covered as a special case by removing the dual steps of FWAL. 

First, we construct the augmented Lagrangian of problem \eqref{eqn:copo-compact} by introducing a dual variable $\vct{y} \in \mathbb{R}^d$ and a penalty parameter $\beta > 0$:
\begin{equation}\label{eqn:augmented-lagrangian}
    L_\beta(\mtx{W};\vct{y}) = \mathrm{Tr}(\mtx{C}\mtx{W}) + \vct{y}^\top (\mathcal{A}\mtx{W} - \vct{v}) + \frac{\beta}{2}\norm{\mathcal{A}\mtx{W} - \vct{v}}^2 \quad \text{for}~\mtx{W}\in\Delta^p.
\end{equation}
The goal is to minimize $L_\beta(\mtx{W};\vct{y})$ with respect to the primal variable $\mtx{W}$ and maximize with respect to the dual variable $\vct{y}$:
\begin{equation}\label{eqn:augmented-lagrangian-problem}
    \min_{\mtx{W} \in \Delta^p} ~ \max_{\vct{y} \in \mathbb{R}^d} ~~ \mathrm{Tr}(\mtx{C}\mtx{W}) +  \vct{y}^\top (\mathcal{A}\mtx{W} - \vct{v}) + \frac{\beta}{2}\norm{\mathcal{A}\mtx{W} - \vct{v}}^2.
\end{equation}
Note, the inner maximization gives an indicator function for $\mathcal{A}\mtx{W}=\vct{v}$:
\begin{equation}
    \max_{y \in \mathbb{R}^m} ~  \vct{y}^\top (\mathcal{A}\mtx{W} - \vct{v}) = 
    \begin{cases} 
    0 & \text{if $\mathcal{A}\mtx{W} = \vct{v}$} \\
    +\infty & \text{otherwise}
    \end{cases}
\end{equation}
Hence, the saddle point problem \eqref{eqn:augmented-lagrangian-problem} is equivalent to our model problem \eqref{eqn:copo-compact}.

The FWAL iteration employs a simple optimization strategy with two main steps, performed on the augmented Lagrangian loss function $L_\beta(\mtx{W};\vct{y})$: \\
(1) A primal step to update $\mtx{W}$, inspired by the FW algorithm, \\
(2) and a dual gradient ascent step to update $\vct{y}$. \\
The penalty parameter, $\beta$, is increased at a specific rate to ensure convergence of $\mtx{W}$ to a feasible solution. 
Next, we describe the algorithm steps in detail. 

\paragraph{Initialization}
Choose an initial penalty parameter $\beta_0 > 0$, and initial primal and dual estimates $\mtx{W}_0 \in \Delta^p$ and $\vct{y}_0 \in \mathbb{R}^d$. In practice, we let $\beta_0 = 1$, and we choose $\mtx{W}$ and $\vct{y}$ as the matrix/vector of zeros. 

At iteration $t = 1,2,\ldots$, we increase the penalty parameter $\beta_t = \beta_0 \sqrt{t+1}$ and perform the following updates:

\paragraph{Primal step} For primal step, we fix the dual variable $\vct{y}_t$ and take a FW step on the primal variable $\mtx{W}_t$ with respect to the augmented Lagrangian loss \eqref{eqn:augmented-lagrangian}. First, we compute the partial derivative of $L_{\beta_t}$ with respect to $\mtx{W}$:
\begin{equation}
    \mtx{G}_t = %
    \mtx{C} + \mathcal{A}^\top \vct{y}_t + \beta_t \mathcal{A}^\top (\mathcal{A}\mtx{W}_t - \vct{v}).
\end{equation}
Then, we find an update direction $\mtx{H}_t \in \Delta^p$ by minimizing the first-order Taylor expansion of $L_{\beta_t}$: %
\begin{equation}
\begin{aligned}
    \mtx{H}_t \in \argmin{\mtx{W} \in \Delta^p} ~ L_{\beta_t}(\mtx{W}_t;\vct{y}_t) + \mathrm{Tr}(\mtx{G}_t(\mtx{W}-\mtx{W}_t))  \equiv \argmin{\mtx{W} \in \Delta^p} ~ \mathrm{Tr}(\mtx{G}_t \mtx{W}).
\end{aligned}
\end{equation}
This step can be written a standard, unconstrained QUBO. Specifically, 
\begin{equation}\label{eqn:LMO-copo}
    \text{if} \quad \vct{w}_t \in \argmin{\vct{w} \in \mathbb{Z}_2^p}~~\vct{w}^\top \mtx{G}_t \vct{w}, \quad \text{then} \quad \mtx{H}_t := \vct{w}_t\vct{w}_t^\top \in \argmin{\mtx{W} \in \Delta^p} ~ \mathrm{Tr}(\mtx{G}_t\mtx{W}). 
\end{equation}
Therefore, we can \textbf{implement and solve this step effectively on an AQC}. This is a key observation for our framework. 

Finally, we update the primal variable $\mtx{W}_t$ by taking a step towards $\mtx{H}_t$:
\begin{equation}
    \mtx{W}_{t+1} = (1-\eta_t) \mtx{W}_t + \eta_t \mtx{H}_t, \quad \text{with step-size}~\eta_t = \frac{2}{t+1}. 
\end{equation}

\paragraph{Dual step}
For dual step, we fix $\mtx{W}_{t+1}$ and take a gradient ascent step on the dual variable with respect to the augmented Lagrangian loss \eqref{eqn:augmented-lagrangian}. The partial derivative of $L_{\beta_t}$ with respect to $\vct{y}$ is
\begin{equation}
    \vct{g}_t = \mathcal{A}\mtx{W}_{t+1} - \vct{v}.
\end{equation}
Then we take a gradient step in this direction
\begin{equation}
    \vct{y}_{t+1} = \vct{y}_t + \gamma_t \vct{g}_t, \quad \text{with step-size $\gamma_t \geq 0$}.
\end{equation}
There are two different strategies for the dual step-size $\gamma_t$, for more details we refer to Section~3.1 in \cite{yurtsever2019conditional}. In practice, we choose a constant step-size $\gamma_t = \beta_0$.

This completes one FWAL iteration. The following proposition, a simple adaptation from \cite[Theorem~3.1]{yurtsever2019conditional}, establishes the convergence rate of FWAL for our model problem \eqref{eqn:copo-compact}. 

\vspace{-0.5em}
\begin{proposition}
Consider FWAL for problem \eqref{eqn:copo-compact}. Choose an initial penalty parameter $\beta_0 > 0$. Assume that the solution set is nonempty, strong duality holds\footnote{Strong duality is a standard assumption for primal-dual methods in optimization.}, and the effective dual domain is bounded (\ie\!, there exists $D<+\infty$ such that $\norm{y_t} \leq D$ at every iteration). Then, the primal sequence $\mtx{W}_t \in \Delta^p$ converges to a solution $\mtx{W}_\star$ with the following bounds on the error:%
\begin{align}
    \mathrm{Tr}(\mtx{C}\mtx{W}_t) - \mathrm{Tr}(\mtx{C}\mtx{W}_\star) & \leq \frac{1}{\sqrt{t}} \left(6\beta_0 p^2 \norm{\mathcal{A}}^2 + \frac{D^2}{2\beta_0}\right) & & \text{(objective suboptimality)}\\
    \norm{\mathcal{A}\mtx{W}_t - \vct{v}} & \leq \frac{1}{\sqrt{t}} \left(2\sqrt{3}p\norm{\mathcal{A}} + \frac{4D}{\beta_0}\right) & & \text{(infeasibility error)}
\end{align}
where $\norm{\mathcal{A}} := \sup \{\norm{\mathcal{A}\mtx{X}}: \norm{\mtx{X}}_F \leq 1\}$ is the operator norm of $\mathcal{A}$.
\end{proposition}

\begin{remark}
We recover FWQP from FWAL by choosing $\vct{y}_0 = \vct{0}$ and $\gamma_t = 0$, in other words, by removing the dual steps. These two methods have similar guarantees with the same rate of convergence up to a constant factor, but FWAL is reported to perform better for most instances in practice \cite{yurtsever2019conditional}. 
\end{remark}

\noindent\textbf{Rounding.} 
We can immediately extract a solution for the original QBO problem \eqref{eqn:qubo} from a solution $\mtx{W}_\star$ of CP reformulation \eqref{eqn:copo-compact}. However, in practice, with finite time and computation, we get only an approximate solution $\hat{\mtx{W}}$. A naive estimate that we extract from $\hat{\mtx{W}}$ can be infeasible for \eqref{eqn:qubo}. To this end, we implement the following rounding procedure: First, we get $\hat{\vct{X}}$ by removing the first row and first column of $\hat{\mtx{W}}$. Next, we compute the best rank-one approximation $\hat{\vct{x}} {\hat{\vct{x}}}^\top$ of $\hat{\mtx{X}}$ with respect to the Frobenius norm.\footnote{This amounts to computing the top singular vector of $\hat{\mtx{X}}$ \cite{mirsky1960symmetric}.}
Finally, as an optional step, we project $\hat{\vct{x}}$ onto the feasible set of \eqref{eqn:qubo}. The set of permutation matrices is the feasible set in our numerical experiments. We use Hungarian algorithm \cite{kuhn1955hungarian} for projection. 

\noindent\textbf{Quantum Annealing (QA).} QA converts a QUBO objective to the equivalent Ising problem that is then solved by a meta-heuristic governed by quantum fluctuations \cite{Farhi2001}. 
Since this analogue optimisation process is prone to different physical disturbances (e.g., state decoherence and cosmic radiation)---and is, hence, non-deterministic---multiple repetitions are required to obtain an optimal solution with high probability. 
Furthermore, the current experimental QA realisations do not easily allow defining high-level constraints; the latter must be integrated so that the QUBO structure is preserved. 
In practice, constraints are formulated as weighted linear terms adjusting qubit couplings and biases \cite{SeelbachBenkner2020,birdal2021quantum}. 
Finding optimal weights (e.g., by a grid search) is a tedious procedure that does not guarantee the generalisation of the selected multipliers across the problems. We provide further details on Quantum annealing in our supplementary material.

\noindent\textbf{On computational complexity.} The convergence of FW is sub-linear and hence may require significant number of iterations, \eg 200-1000. 
At each iteration, Q-FW involves attempting to solve an NP-Hard QUBO problem whose computational complexity class is $\mathrm{FP}^{\mathrm{NP}}$-complete\footnote{A binary relation $P(x,y)$, 
is in $\mathrm{FP}^{\mathrm{NP}}$ if and only if there is a deterministic polynomial time algorithm that can determine whether $P(x,y)$ holds given both $x$ and $y$.}~\cite{yasuoka2021computational}.
Thanks to the exploitation of quantum phenomena, QA can bring a quadratic improvement reducing the theoretical complexity from $O(e^{{N}})$ to $O(e^{\sqrt{N}})$, in a similar vein to Grover algorithm~\cite{albash2018adiabatic,grover1996fast}. Though, it is not straightforward to get a problem-specific, realistic estimate of the time complexity of the QA  process. 
Nevertheless, fixing a constant annealing time and a constant number of repetitions, as we do for our small problems, can lead to an optimistic, polynomial time algorithm~\cite{arthur2021qubo}. 

\vspace{-2mm}\section{Experimental Evaluation}\label{sec:exp}\vspace{-3mm}
The proposed approach (Q-FWAL) is general and not tailored towards a specific problem. Hence, we assess its validity
in realizing quantum versions two different problems, \emph{graph matching} and \emph{permutation synchronization}, both requiring equality constraints to be accounted for.\footnote{While still providing a way to handle inequalities in our supplementary material, we leave it as a future work to study problems with inequality constraints.} We use problem-specific synthetic and real datasets to showcase the effectiveness of our approach. 

\paragraph{Implementation details} In both of the experiments we use the DWave Advantage 4.1 system~\cite{mcgeoch2021advantage} which has at least 5,000 qubits and $\sim$35,000 couplers. 
Except the ablation studies, we use 50 or 250 annealing cycles of $20{\mu}s$ in each iteration with an annealing schedule of $100{\mu}s$ breaks. We set the chain strength $\xi$ according to the \emph{maximum chain strength criterion}: We inspect the minor embedding calculated by Cai \textit{et al.}~\cite{Cai2014} and set $\xi = s_{\text{max}} + \omega$, with $s_{\text{max}}$ being the maximum chain length in the minor embedding and $\omega = 0.5$ is the strengthening weight. 
If we observe frequent chain breaks for larger  problems, we increase $\omega$ to $3.0$. We access DWave at each iteration through the \emph{Leap2} API~\cite{DWave_Leap}.
We investigate three modes of Q-FW: (i) with intermediary exhaustive solution instead of DWave (FWAL), (ii) without Hungarian rounding (Q-FWAL relaxed) and (iii) the full configuration (Q-FWAL). Note that vanilla FWAL (i) cannot be applied to large problems due to the combinatorial explosion. 
In all of our problems, we are interested in linear permutation constraints, as those are the most common in CV problems. Hence, we use \emph{Hungarian algorithm}~\cite{kuhn1955hungarian} as the projector onto the constraint set (\cf~\emph{Rounding} in \S\ref{sec:QFW}) and formulate permutation-ness into linear constraints as in~\cite{birdal2021quantum,SeelbachBenkner2020} (\cf supplementary material).

\vspace{-2mm}
\subsection{Quantum Graph Matching (QGM)}\label{sec:QGM}
In general, 3D vision problems relate two abstract shape/image manifolds $\Man_1$ and $\Man_2$. In many applications, these manifolds can be sampled by two point clouds (e.g. keypoints) $\Xset_1\in\R^{\texttt{N}_1\times \texttt{n}}$ and $\Xset_2\in\R^{\texttt{N}_2\times \texttt{n}}$ where $\texttt{n}$ is the dimensionality of the problem domain, \eg two for images, three for meshes and etc. We further assume a distance function $\phi(.)$ defined over the points of these point clouds. The quadratic assignment problem (QAP) then takes the form:
\begin{equation}
    \max_{\mtx{\Pi}}\,\,\vec(\mtx{\Pi})^\top \,\mtx{Q}_{\mathrm{QGM}}\, \vec(\mtx{\Pi}) \quad \mathrm{subject~to } \quad  \mtx{\Pi}\in\Pset
\end{equation}
where $\mathcal{P}$ denotes the set of (partial) permutations and $\vec(\cdot)$ acts as a vectorizer. Assuming $\texttt{N}:=\texttt{N}_1=\texttt{N}_2$, \ie \emph{total} permutations, $\mtx{Q}_{\mathrm{QGM}}\in\R^{\texttt{N}^2\times \texttt{N}^2}$ denotes a \emph{ground cost} matrix or the quadratic energy measuring the gain of matching $\Man_1$ and $\Man_2$ by a sub-permutation $\mtx{\Pi}$, computed using the distance $\phi(\cdot)$. %
\begin{table}[tb]
  \centering
  \caption{Evaluations of graph matching on random problem instances with different sizes \cite{SeelbachBenkner2020}. We report mean normalized energies over ten instances (the lower the better). Last five columns correspond to the variants of our method.}
   \setlength\tabcolsep{6.5pt}
    \begin{tabular}{lcccc|ccccc}
          & \cite{SeelbachBenkner2020}  & \cite{SeelbachBenkner2020}  & \cite{Bernard_2018_CVPR}   & \cite{Kirkpatrick1983}   &       & Q-FWAL & Q-FWAL & \multicolumn{1}{c}{Q-FWAL} & Q-FWAL \\
    $\texttt{N}$     & ins. & row. & DS*   & SA    & FWAL  & relaxed (50) & (50)  & \multicolumn{1}{c}{relaxed (250)} & (250) \\
    \midrule
    3     & 1.49  & 2.12  & 0.85  & 0.82  & \textbf{7e-4} & 1.72  & 0.093 & 1.72  & \textbf{7e-4} \\
    4     & 5.68  & 7.37  & 0.43  & 2.43  & \textbf{1.3e-3} & 3.41  & 1.82  & 0.23  & \textbf{1.43e-3} \\
    \end{tabular}%
  \label{tab:QGM}\vspace{-4mm}
\end{table}%

\paragraph{Baselines \& dataset} We benchmark QGM against the exhaustive solution, obtained by searching over all possible permutations, as well as against the first AQC approach which was proposed by Benkner~\etal~\cite{SeelbachBenkner2020} who used multiple strategies (\eg \emph{inserted}, \emph{row-wise}) to inject soft-permutation constraints into QUBO. This required tuning of a parameter $\lambda\in\R$, whose large values are found to cause problems~\cite{birdal2021quantum,SeelbachBenkner2020,zaech2022adiabatic}. As a heuristic,~\cite{SeelbachBenkner2020} suggested a \emph{spectral-gap}\footnote{the difference between the lowest and second-lowest energy state / eigen-value} analysis to bound the regularization coefficient $\lambda$. 
We also include: (i) the result obtained by running \emph{simulated annealing} (SA) \cite{Kirkpatrick1983} on a CPU (the  implementation from the Ocean  tools~\cite{DWave_Leap}); (ii) a state of the art classical graph matching algorithm~\cite{Bernard_2018_CVPR}.

To assess, we use two sets of ten random problem instances with $\texttt{N}=3$ and $\texttt{N}=4$ as in \S 5.1 of~\cite{SeelbachBenkner2020}. 
The ground-truth permutations are calculated by brute force and compared qualitatively with the expected outcomes on real data.
The number or qubits in the minor embeddings equals to $14$ ($\texttt{N}=3$) and $40$ ($\texttt{N}=4$). 

\paragraph{Results} We report the \emph{mean normalized energies over ten instances} in~\cref{tab:QGM}. 
This quantity is obtained by first shifting all energies by the minimum energy (of the ground-truth solution) and then averaging them. Clearly, FWAL and Q-FWAL perform the best on this experiment. However, FWAL cannot be scaled to large problems, and as we will see later in \S\ref{sec:QPS}, Q-FWAL is able to handle much larger problems thanks to the advances in AQCs. DS* is a powerful classical algorithm, yet it cannot match the errors we achieve. SA is good for small problems, but its solution quality quickly drops with the problem size. Finally, it is visible that 50 cycles might be insufficient to get high quality results. %

\begin{figure}[bht]\vspace{-2mm}
     \centering
         \includegraphics[width=\textwidth]{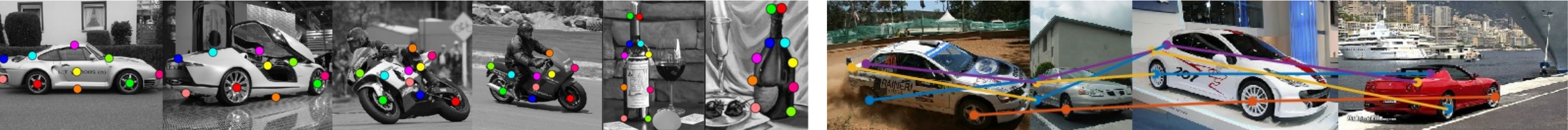}
        \caption{Willow Dataset~\cite{cho2013learning}. (\textbf{left}) Manual annotations of keypoints. (\textbf{right}) Ground truth multi-image matches.}
        \label{fig:willow}\vspace{-3mm}
\end{figure}
\vspace{-5mm}\subsection{Quantum Permutation Synchronization (QPS)} \vspace{-3mm}
\label{sec:QPS}
Many multi-shape/view/instance computer vision problems can be solved by synchronization, including shape (point set) alignment~\cite{gao2021isometric,huang2021multiway}, structure from motion~\cite{birdal2018bayesian,birdal2020synchronizing,govindu2004lie}, multi-view matching~\cite{maset2017,birdal2019probabilistic}, point cloud registration~\cite{gojcic2020learning,huang2019learning} and motion segmentation~\cite{huang2021multibodysync,arrigoni2022multi}. %

A specific branch, \emph{permutation synchronization} seeks to find globally consistent image/shape matches from a set of relative matches over a collection. In particular, consider a \emph{collection} of $\texttt{K}$ point sets $\Xset_1,\dots,\Xset_\texttt{K}$ \footnote{Such sets are easy to obtain by keypoint detection or sampling either on images or on shapes, \eg by detecting $N$ landmarks per image, in a $M$-view image collection.} of $\texttt{N}$ points each such that there exists a bijective map for each pair $(\Xset_i,\Xset_j)$. We assume the availability of a set of noisy \emph{relative} permutations $\{\Pmat_{ij}:\Xset_i\to\Xset_j\}_{ij}$ estimated in \emph{isolation}, \ie independently. Our goal is then to solve this multi-graph matching problem even when a significant fraction of the pairwise matches are incorrect.
To this end, a large body works minimize a \emph{cycle-consistency loss}, that is shown to be equivalent to a QUBO (\cf~\cite{birdal2021quantum} for a proof):
\begin{align}
    \argmin{\{\X_i \in \Pset_\texttt{n}\}} \sum_{(i,j)\in \Edge} \|\Pmat_{ij} - \X_i \X_j^\top \|^2_\mathrm{F}=\argmin{\{\X_i \in \Pset_\texttt{n}\}} \, \x^\top \Q_{QPS} \x.
\end{align}
Here, $\x=[\cdots \x_i^\top \cdots]^\top$ and $\x_i=\vec(\X_i)$ depict the \emph{canonical ordering} of points. The first AQC approach to this problem is proposed by Birdal~\etal~\cite{birdal2021quantum}, who, similar to~\cite{SeelbachBenkner2020}, regularize $\Q_{QPS}$ to incorporate permutation-ness as a soft constraint. Note that, this whole problem has a gauge freedom, where we can freely choose $\X_0$ \eg, as an identity matrix (\cf~supplementary material). %

\paragraph{Datasets}
As a real dataset, we follow~\cite{birdal2021quantum} and use the kindly provided subset of the Willow Object Classes~\cite{cho2013learning} composed of four categories (\emph{duck}, \emph{car}, \emph{winebottle}, \emph{motorbike}) with 40 RGB images each, acquired \emph{in the wild} (\cf~\cref{fig:willow}). This subset contains multiple sets of four points sampled out of ten annotations. This leads to 35 small problems per category each of which is a fully connected graph of all four consecutive frames. Initial permutations are obtained via a Hungarian algorithm~\cite{munkres1957algorithms} applied to matching costs obtained by  Alexnet~\cite{krizhevsky2012imagenet} features. As the data is manually annotated, the ground-truth relative maps are known.

\begin{table}[t]
  \centering
  \caption{Evaluations on Willow Dataset.}
  \setlength{\tabcolsep}{3pt}
  \resizebox{\columnwidth}{!}
  {
    \begin{tabular}{lcccc|c}
          & Car   & Duck  & Motorbike & Winebottle & Average \\
\cmidrule{2-6}   
    MatchEIG \cite{maset2017}   & {0.81  $\pm$  0.083} & {0.86  $\pm$  0.102} & {0.77  $\pm$  0.059} & {0.87  $\pm$  0.107} & {0.83  $\pm$  0.088} \\
    MatchALSS \cite{zhou2015multi}   & {{0.84  $\pm$  0.095}} & 0.90  $\pm$ 0.102 & 0.81  $\pm$  0.078 & 0.94  $\pm$  0.092 & 0.87  $\pm$  0.092 \\
    MatchLIFT \cite{huang2013consistent}  & {{0.84  $\pm$  0.102}} & 0.90  $\pm$  0.103 & 0.81  $\pm$  0.078 & 0.94  $\pm$  0.092 & 0.87  $\pm$  0.094 \\
    MatchBirkhoff \cite{birdal2019probabilistic} & {0.84  $\pm$  0.094} & 0.90  $\pm$  0.107 & 0.81  $\pm$  0.079 & 0.94  $\pm$  0.093 & 0.87  $\pm$  0.093 \\
    QuantumSync \cite{birdal2021quantum} & {{0.84  $\pm$  0.104}} & 0.90  $\pm$  0.104 & 0.81  $\pm$  0.080 & {{0.93  $\pm$  0.095}} & 0.87  $\pm$  0.096 \\
    \cite{birdal2021quantum}-search & {{0.84  $\pm$  0.104}} & {0.91  $\pm$  0.115} & {0.82  $\pm$  0.10} & {0.95  $\pm$  0.096} & {0.88  $\pm$  0.104} \\
    \midrule
    Q-FWAL (ours) & {\textbf{0.92  $\pm$  0.094}} & {\textbf{0.97  $\pm$  0.072}} & {\textbf{0.89 $\pm$  0.093}} & {\textbf{0.99  $\pm$  0.044}} & {\textbf{0.94  $\pm$  0.076}} \\
    \end{tabular}%
    }
  \label{tab:willow}\vspace{-3mm}
\end{table}%

\paragraph{Baselines} We compare \methodname~against the classical algorithms of MatchEIG \cite{maset2017}, MatchALS~\cite{zhou2015multi}, MatchLift~\cite{huang2013consistent},  MatchBirkhoff~\cite{birdal2019probabilistic} as well as the first Quantum approach, QuantumSync~\cite{birdal2021quantum}. QuantumSync uses $\lambda=2.5$ in all experiments. The \emph{exhaustive} solution is obtained by enumerating all possible permutations. Note that due to the limitations in the available 
DWave time, we had to implement an \emph{early-stopping} heuristic whose details are provided in the supplementary document.
The number or qubits in the minor embeddings in this experiment (for $\Q\in\R^{64\times 64}$) was ${\approx}270$, and the chain length did not exceed eight.

\paragraph{Results} We follow the protocol of Birdal \textit{et  al.}~\cite{birdal2021quantum} and report  in~\cref{tab:willow}, the portion of correct bits \ie \emph{accuracy}. 
Our approach consistently and significantly outperforms both the classical algorithms and the state-of-the-art quantum approach,  QuantumSync~\cite{birdal2021quantum}. \emph{\cite{birdal2021quantum}-search} denotes the softly-constrained search detailed in~\cite{birdal2021quantum}. Overall, Q-FW is more applicable to problems of growing size. 

\vspace{-3mm}\subsection{Ablation Studies} \vspace{-2mm}
\begin{wrapfigure}{r}{0.32\textwidth}
\vspace{-8mm}
\includegraphics[width=0.32\textwidth]{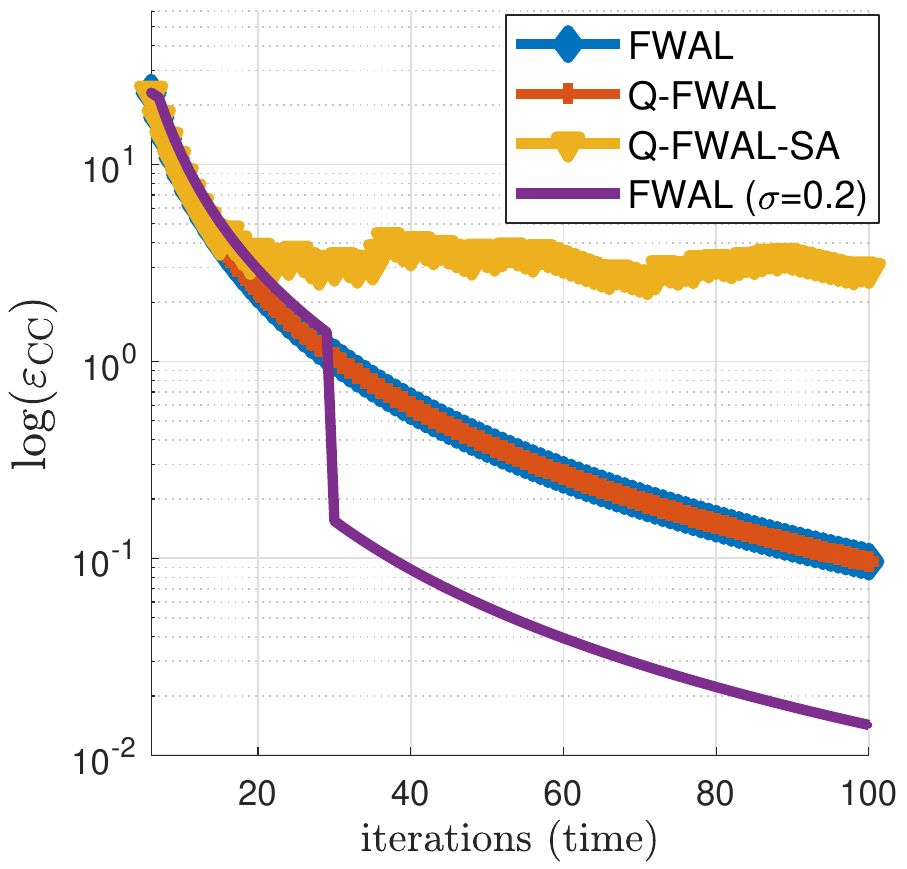}
\vspace{-12mm}
    \label{fig:tight}
\end{wrapfigure}
\textbf{Tightness of the copositive relaxation.} To assess the tightness of our algorithm, we randomly generate fully connected, synthetic synchronization problems with $\texttt{N}=3$ and $\texttt{K}=3$ with different noise levels $\sigma\in\{0,0.2\}$. For this small problem we could use an exact QUBO solver and monitor the convergence of the relaxed problem to the ground truth (GT):
$\varepsilon_\mathrm{CC}=|\mathrm{Tr}(\Q\X_t) - \mathrm{Tr}(\Q\X^{\mathrm{gt}}_t)|$ where $\X^{\mathrm{gt}}$ is obtained by lifting the GT permutations. As shown on the right, $\varepsilon_\mathrm{CC}$ decreases monotonically for all methods, even in the case of noise. Moreover, our D-Wave implementation strictly matches FWAL.

\begin{figure}[t!]
     \centering
     \begin{subfigure}[b]{0.325\textwidth}
         \centering
         \includegraphics[width=\textwidth]{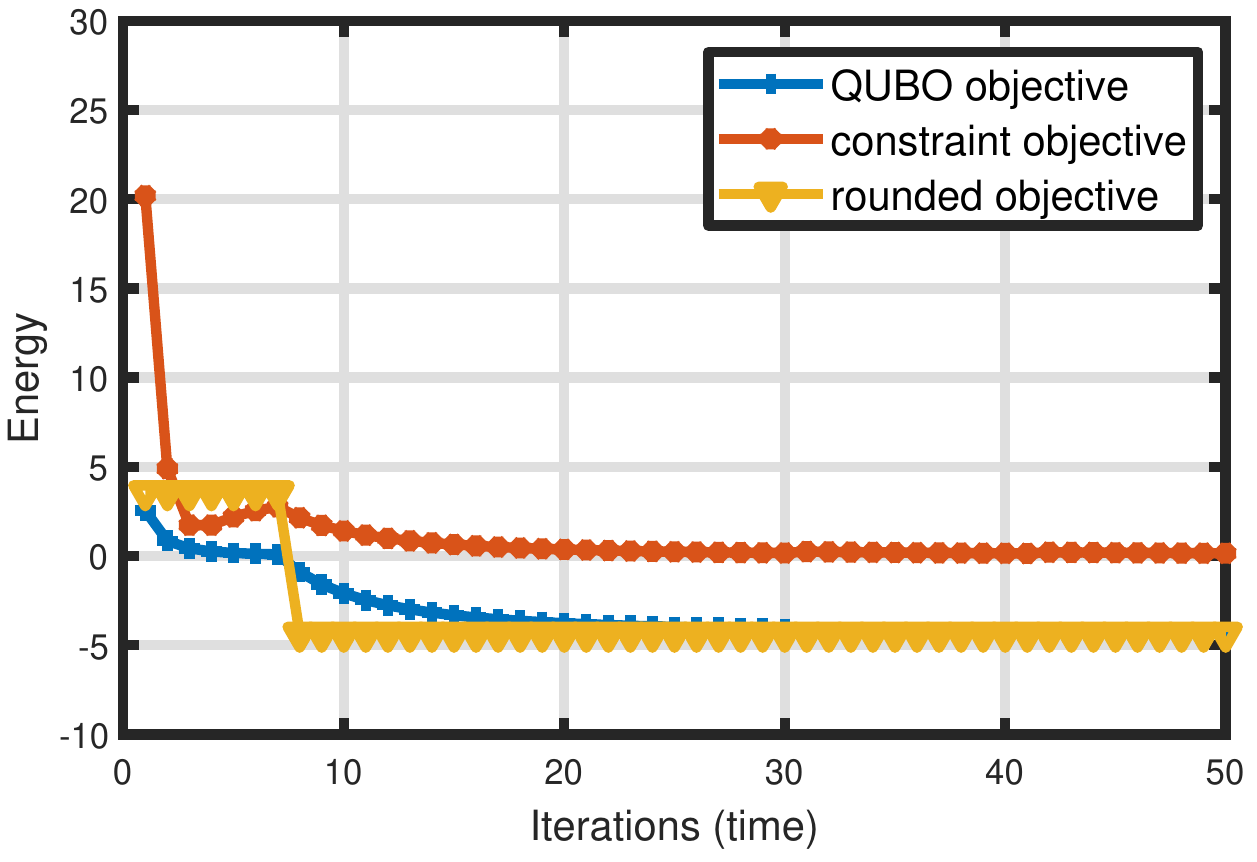}
         \caption{\footnotesize Graph matching\\ ($\Q\in\R^{9\times 9}$)}
         \label{fig:qgmiter9}
     \end{subfigure}
     \hfill
     \begin{subfigure}[b]{0.325\textwidth}
         \centering
         \includegraphics[width=\textwidth]{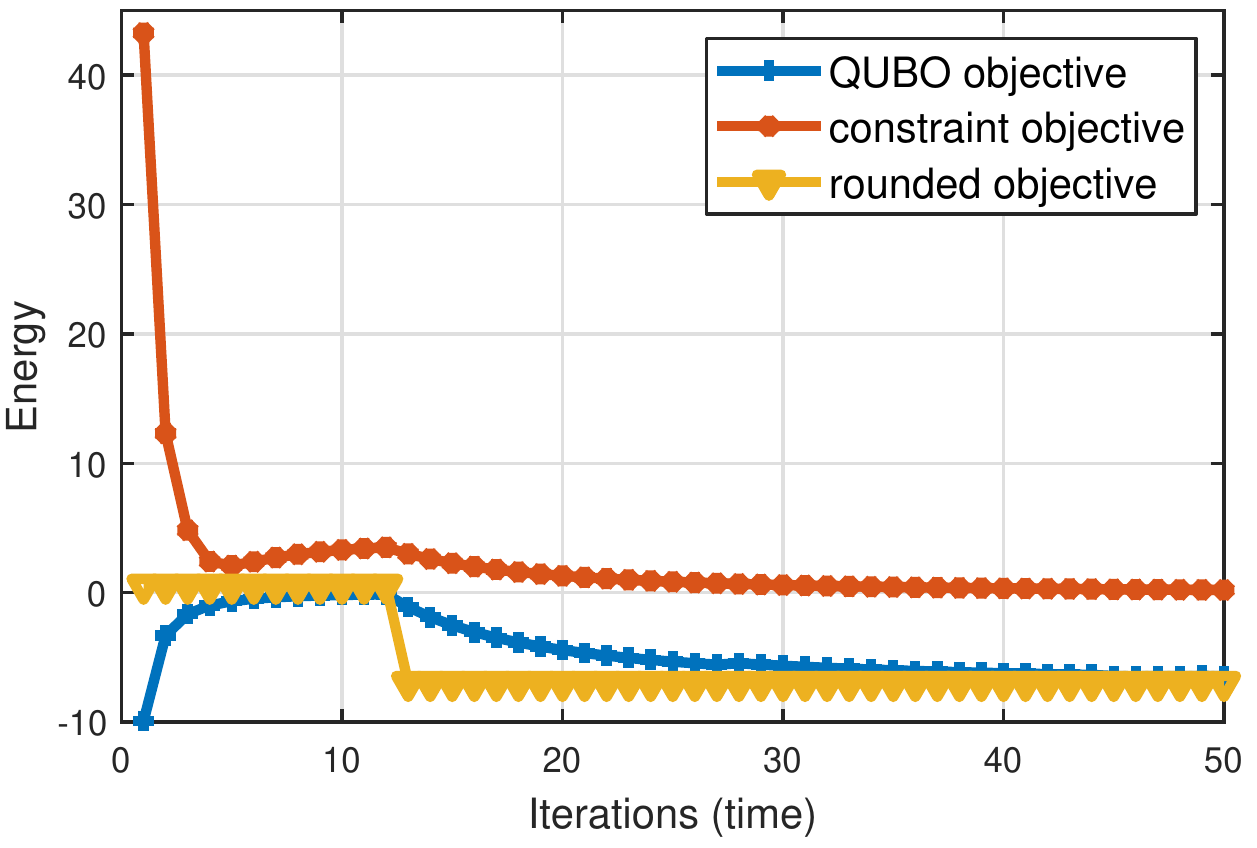}
         \caption{\small Graph matching\\ ($\Q\in\R^{16\times 16}$)}
         \label{fig:qgmiter16}
     \end{subfigure}
     \hfill
     \begin{subfigure}[b]{0.325\textwidth}
         \centering
         \includegraphics[width=\textwidth]{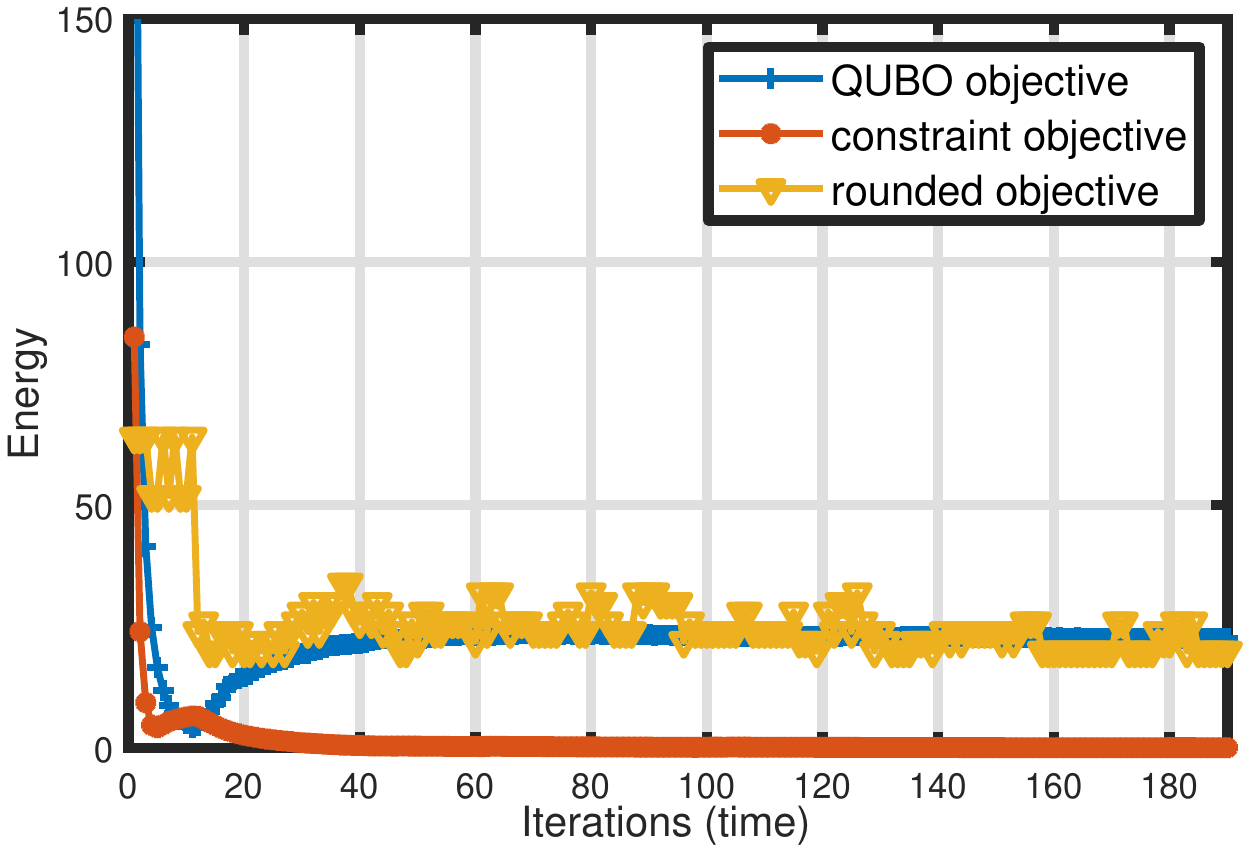}
         \caption{\small Synchronization\\ ($\Q\in\R^{64\times 64}$).}
         \label{fig:synciter}
     \end{subfigure}
        \caption{Solving two graph matching and one synchronization problem using Q-FWAL. The problem gets more complex from left to right. Thus, the required number of iterations to converge increases.\vspace{-4mm}}
        \label{fig:iterations}
\end{figure}

\paragraph{Monitoring convergence}
As heuristic fearly stopping criteria are harmful for the convergence guarantees we provide, it is of interest to see how our algorithm behaves as iterations progress. In~\cref{fig:iterations} we plot minimization curves for different problems we consider: two graph matching (a,b) and one synchronization (c). For each problem, we plot the QUBO objective, infeasibility eror (constraint objective) and the error attained after Hungarian rounding. It is visible that simplicity of the problem has a positive impact on finding good solutions early on. For larger problems, settling on a good solution can take $>200$ iterations, when early stopping is not used. We also note that the QUBO ojective converges to the rounded objective, indicating the tightness of our relaxation.

\begin{figure}[ht]
     \centering
         \centering
         \includegraphics[width=\textwidth]{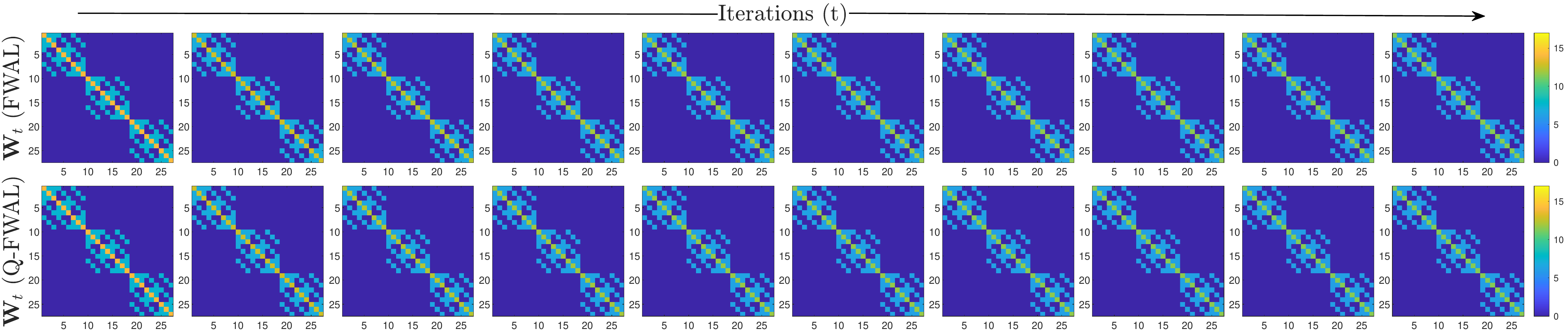}
         \caption{Evolution of the gradient $\W_t$ for $0<t<100$ sampled in steps of $10$: FWAL (top) and Q-FWAL (bottom).}
         \label{fig:Qs}\vspace{-5mm}
\end{figure}

\paragraph{On the evolution of sub-problems \& sparsity}
We now visually compare the sub-problems emerging in solving the noiseless, synthetic synchronization problem (detailed in the previous experiment and in our supplementary), for our exact method and for the D-Wave implementation. As seen in~\cref{fig:Qs}, there is no noticeable difference between the two evolutions, confirming that D-Wave could solve the sub-QUBO-problems reliably. Moreover, over iterations the sparsity pattern of $\W_t$ is fixed, which means that we could compute the minor embedding\footnote{requires solving a combinatorial optimization problem with heuristics.}, and re-use it throughout Q-FW. This ability of avoiding repetitive minor embeddings makes Q-FW a practically feasible algorithm.\vspace{-2mm}

\vspace{-5mm}\section{Discussions and Conclusion }\label{sec:conc}\vspace{-3mm}
We have proposed Q-FW, a quantum computation backed, hybrid Frank Wolfe Augmented Lagrangian method. 
Thanks to the tight copositive relaxation and the QUBO formulation, our algorithm has successfully satisfied linear (in)equality constraints, such as permutation-ness, arising in many computer vision applications. We have solved the intermediary QUBO problems on a quantum computer to obtain high quality update directions and demonstrated the validity of Q-FW both on graph matching and on permutation synchronization.

\noindent\textbf{Limitations.} The most obvious concern is the sub-linear convergence of our algorithm, which could sometimes require a large number of iterations. However, we observed in practice a maximum of 300-400 iterations can be sufficient thanks to the good quality of the DWave solver. We are also limited by the small problem sizes, just likes the previous studies~\cite{birdal2021quantum,SeelbachBenkner2020}. Yet, quantum computers evolve steadily and we are hopeful that the problems we could solve will only grow with time.

\noindent\textbf{Looking forward.} Q-FWAL leaves ample room for future works. First, a plethora of QCV algorithms concerned with constraint satisfaction can benefit our approach. Using our algorithm to ensure constraints other than permutations (especially inequalities like partial permutations) is a future study. We also like to deploy our algorithm in training vector-output RELU networks~\cite{sahiner2020vector}. 
\section*{Acknowledgements}

A.Y. received support from the Wallenberg AI, Autonomous Systems and Software Program (WASP) funded by the Knut and Alice Wallenberg Foundation.

\bibliographystyle{splncs04}
\bibliography{bibliography_long,biblio}

\begin{appendices}

\setcounter{equation}{0}
\renewcommand{\theequation}{S.\arabic{equation}}

\section{Theoretical Aspects \& Discussions}\label{sec:suppTheory}
\vspace{-2mm}
\subsection{Permutation-ness as a Linear Constraint}\vspace{-1mm}
The formulation of permutation-ness into linear constraints appeared both in QGM~\cite{SeelbachBenkner2020} and in QuantumSync~\cite{birdal2021quantum}. We include a brief description here for completeness.
A \textit{permutation matrix} is defined as a sparse, square binary matrix, where each column or row contains only a single non-zero entry:
\begin{equation}
\label{eq:def:perm}
\mathcal{P}_n := \{\Pmat \in \{0,1\}^{n\times n} : \Pmat \one_n = \one_n\,,\,\one_n^\top \Pmat = \one_n^\top\}.
\end{equation}
where $\one_n$ denotes a $n$-dimensional ones vector. Every $\Pmat \in \mathcal{P}_n$ is a \textit{total} permutation matrix and $P_{ij}=1$ implies that point $i$ is mapped to element $j$. Note, $\Pmat^\top=\Pmat^{-1}$.

During optimization, permutation-ness could be imposed on a binary vector/matrix by introducing linear constraints $\A\x=\vct{b}$: rows and columns sum to one as in~\cref{eq:def:perm}. Given $\x_i=\vec(\X_i)$, this amounts to having $\bb_i=\one$ and 
\begin{equation} 
    \A_i=\begin{bmatrix}
    \Id \otimes \one^\top \\
    \one^\top \otimes \Id
    \end{bmatrix}.
\end{equation}
Put simply, the matrix $\A_i$ is assembled as follows: in row $j$ with $1\leq j \leq n$, the ones are placed in columns $(j-1)\cdot n + 1 $ to $(j )\cdot n $. In a row $j$ with $ j>n$, ones will be placed at $ (j-n )+ p\cdot n $ for $p \in \{ 0,...,n-1 \} $. To enforce the permutation-ness of all the individual $\x_i$ that make up $\x\in\R^{n^2\times m}$, we construct a $n^2 \times 2n$ block-diagonal matrix $\A = \mathrm{diag}(\A_1, \A_2, \dots, \A_m)$. 

    \begin{equation}
    \A_i= \begin{bmatrix}
1 &1&0&0 \\
0&0&1&1 \\
1&0&1&0 \\
0&1&0&1
    \end{bmatrix}, \quad \bb_i = \one =  \begin{bmatrix}
    1\\
    1\\
    1\\
    1
    \end{bmatrix}.
\end{equation}
\vspace{-5mm}
\subsection{On Permutation Synchronization \& Gauge Freedom} \vspace{-2mm}
A close look to the presented permutation synchronization problem reveals that it is \textbf{non-convex} in $(\Pmat_i, \Pmat_j)$, but \textbf{convex when} one odd, \eg $\Pmat_i$, is fixed during optimization. In fact, if $\Pmat_i$ is considered to be fixed, this problem resembles a matrix averaging under the metric the Frobenius norm. 

The formulation in \S 5.2 is subject to a freedom in the choice of the reference or the \emph{gauge}~\cite{belov2019geometry,chatterjee2013efficient,birdal2021quantum}. In other words, the solution set can be transformed arbitrarily by a common $\Pmat_g$, still satisfying the consistency constraint:
\begin{align}
    E(\{\X_i\Pmat_g\}) &= \sum_{(i,j)\in \Edge} \|\Pmat_{ij} - (\X_i\Pmat_g) (\X_j\Pmat_g)^\top \|^2_\mathrm{F}\\
    &= \sum_{(i,j)\in \Edge} \|\Pmat_{ij} - \X_i\Pmat_g \Pmat_g^\top\X_j^\top \|^2_\mathrm{F}\\
    &= \sum_{(i,j)\in \Edge} \|\Pmat_{ij} - \X_i\X_j^\top \|^2_\mathrm{F}\\
    &= E(\{\X_i\}).
\end{align}
The last equality follows from the orthogonality of permutation matrices. 
In practice, a gauge can be fixed by setting one of the vertex labels to identity: $\X_1 = \Id$. However, for convenience, we do not explicitly account for gauge freedom. We transform the first node to identity, only after obtaining the full solution.

\subsection{Extended Notation}
Q-FW involves a lifting procedure that maps a QBO problem with $n$ variables and $m$ equality constraints into a copositive program with $(n+1)^2$ variables and $2m+n+1$ equality constraints. This dimensionality expansion complicates the notation. %
For the ease of presentation, we introduce a compact notation in (\S3) of the main text. Here, we revisit this notation for clarity.

First, we define the primal and dual dimensions $p = n+1$ and $d = 2m+n+1$. Then, our primal variable is a $p \times p$ completely positive matrix $\mtx{W} \in \Delta^p$, and our dual variable $\vct{y} \in \mathbb{R}^d$. $\mtx{W}$ relates to $\vct{x} \in \mathbb{R}^n$ and $\mtx{X} \in \mathbb{R}^{n\times n}$ by
\begin{equation}
\mtx{W} = \begin{bmatrix} 
W_{11} & ~~ \vct{x}^\top \\ 
\vct{x} & ~ \mtx{X}
\end{bmatrix},
~ \text{with the constraint $W_{11}=1$}.
\end{equation}

Then, we define a linear map $\mathcal{A}:\mathbb{R}^{p \times p} \to \mathbb{R}^d$ and $\vct{v} \in \mathbb{R}^d$ to simplify the writing of the constraints as $\mathcal{A}\mtx{W} = \vct{v}$. Explicitly, $\mathcal{A}$ and $\vct{v}$ are defined by
\begin{equation}\label{eqn:Aoperator}
    \underbrace{\begin{bmatrix} 
    W_{11} \\
    X_{11} - x_1\\
    \vdots \\
    X_{nn} - x_n \\
    \vct{a}_1^\top \vct{x} \\
    \vdots \\
    \vct{a}_m^\top \vct{x} \\
    \mathrm{Tr}(\mtx{A}_1\vct{X}) \\
    \vdots \\
    \mathrm{Tr}(\mtx{A}_m\vct{X})
    \end{bmatrix}}_{\displaystyle\mathcal{A}\mtx{W}} = 
    \underbrace{\begin{bmatrix} 
    1 \\
    0\\
    \vdots \\
    0 \\
    b_1 \\
    \vdots \\
    b_m \\
    b_1^2 \\
    \vdots \\
    b_m^2
    \end{bmatrix}}_{\displaystyle\vct{v}} 
\end{equation}

\subsection{Convergence Analysis}
Our Q-FWAL algorithm is a special instance of Frank Wolfe with Augmented Lagrangian (FWAL) methods where Q-FWAL uses the QUBO-specific D-Wave solver and a tight copositive relaxation.  With the observation that copositive relaxation does not have an effect on the convergence of FWAL and the assumption that the solver is \emph{exact}\footnote{Although we do know currently that this is not true, quantum revolution might enable computers, which largely satisfy this assumption in the future.}, it is possible to consult the FW literature for a convergence analysis. We now present the proof of convergence rate of FWAL (see Proposition~1 in the main text) for completeness. The original proof appears in \cite{yurtsever2019conditional}. Our presentation closely follows the exposition in  \cite[Section~SM1.6]{yurtsever2021scalable}. 

First, we exploit smoothness of $L_{\beta_t}$ in the primal argument:
\begin{equation}\label{eqn:proof-smoothnes}
\begin{aligned}
    L_{\beta_t}(\mtx{W}_{t+1},\vct{y}_t) 
    & \leq L_{\beta_t}(\mtx{W}_t,\vct{y}_t) + \mathrm{Tr}(\mtx{G}_t (\mtx{W}_{t+1} - \mtx{W}_t)) + \frac{1}{2} \beta_t \norm{\mathcal{A}(\mtx{W}_{t+1} - \mtx{W}_t)}_F^2 \\
    & = L_{\beta_t}(\mtx{W}_t,\vct{y}_t) + \eta_t \mathrm{Tr}(\mtx{G}_t (\mtx{H}_t - \mtx{W}_t)) + \frac{1}{2} \beta_t \eta_t^2 \norm{\mathcal{A}(\mtx{H}_t - \mtx{W}_t)}_F^2 \\
    & \leq L_{\beta_t}(\mtx{W}_t,\vct{y}_t) + \eta_t \mathrm{Tr}(\mtx{G}_t (\mtx{H}_t - \mtx{W}_t)) + \frac{1}{2} \beta_t \eta_t^2 \norm{\mathcal{A}}^2 p^2 \\
    & \leq L_{\beta_t}(\mtx{W}_t,\vct{y}_t) + \eta_t \mathrm{Tr}(\mtx{G}_t (\mtx{W}_\star - \mtx{W}_t)) + \frac{1}{2} \beta_t \eta_t^2 \norm{\mathcal{A}}^2 p^2. 
\end{aligned}
\end{equation}
The second line follows by definition of $\mtx{W}_{t+1}$, the third line holds because the Frobenius-norm diameter of $\Delta^p$ is $p$, and the last line depends on the fact that $\mtx{H}_t$ minimizes $\mathrm{Tr}(\mtx{G}_t \; \cdot \;)$. 

Next, we use the definition of $\mtx{G}_t$ to bound \begin{equation}\label{eqn:proof-lmo}
\begin{aligned}
    \mathrm{Tr}(\mtx{G}_t(\mtx{W}_\star - \mtx{W}_t)) 
    & = \mathrm{Tr}\Big(\big(\mtx{C} + \mathcal{A}^\top\vct{y}_t + \beta_t \mathcal{A}^\top(\mathcal{A}\mtx{W}_t - \vct{v})\big)\big(\mtx{W}_\star - \mtx{W}_t\big)\Big) \\
    & \hspace{-2em} = \mathrm{Tr}(\mtx{C}(\mtx{W}_\star - \mtx{W}_t)) + \big(\vct{y}_t + \beta_t (\mathcal{A}\mtx{W}_t - \vct{v})\big)^\top \big(\mathcal{A}\mtx{W}_\star - \mathcal{A}\mtx{W}_t\big) \\
    & \hspace{-2em} = \mathrm{Tr}(\mtx{C}(\mtx{W}_\star - \mtx{W}_t)) + \big(\vct{y}_t + \beta_t (\mathcal{A}\mtx{W}_t - \vct{v})\big)^\top \big(\vct{v} - \mathcal{A}\mtx{W}_t\big) \\
    & \hspace{-2em} = \mathrm{Tr}(\mtx{C}\mtx{W}_\star) - L_{\beta_t}(\mtx{W}_t,\vct{y}_t) - \frac{\beta_t}{2}\norm{\mathcal{A}\mtx{W}_t - \vct{v}}^2 \\
\end{aligned}
\end{equation}
where we used the fact that $\mathcal{A}\mtx{W}_\star = \vct{v}$. 

We combine \eqref{eqn:proof-smoothnes} with \eqref{eqn:proof-lmo} and subtract $\mathrm{Tr}(\mtx{C}\mtx{W}_\star)$ to get
\begin{equation}
\begin{multlined}
    L_{\beta_t}(\mtx{W}_{t+1},\vct{y}_t) - \mathrm{Tr}(\mtx{C}\mtx{W}_\star)
    \leq (1-\eta_t) \Big( L_{\beta_t}(\mtx{W}_t,\vct{y}_t) - \mathrm{Tr}(\mtx{C}\mtx{W}_\star) \Big) \\
    - \frac{1}{2}\beta_t\eta_t\norm{\mathcal{A}\mtx{W}_t - \vct{v}}^2+ \frac{1}{2} \beta_t \eta_t^2 \norm{\mathcal{A}}^2 p^2. 
\end{multlined}
\end{equation}
Now, we update the penalty parameter on the right hand side,
\begin{equation}\label{eqn:proof-beta-update}
\begin{multlined}
    L_{\beta_t}(\mtx{W}_{t+1},\vct{y}_t) - \mathrm{Tr}(\mtx{C}\mtx{W}_\star)
    \leq (1-\eta_t) \Big( L_{\beta_{t-1}}(\mtx{W}_t,\vct{y}_t) - \mathrm{Tr}(\mtx{C}\mtx{W}_\star) \Big) \\
    \begin{aligned}
    & + \frac{1}{2}(1-\eta_t)(\beta_t - \beta_{t-1})\norm{\mathcal{A}\mtx{W}_t - \vct{v}}^2\\
    & ~- \frac{1}{2}\beta_t\eta_t\norm{\mathcal{A}\mtx{W}_t - \vct{v}}^2+ \frac{1}{2} \beta_t \eta_t^2 \norm{\mathcal{A}}^2 p^2. 
    \end{aligned}
\end{multlined}
\end{equation}
By design, our parameter choices for $\eta_t$ and $\beta_t$ ensures that
\begin{equation}
    (1-\eta_t)(\beta_t - \beta_{t-1}) \leq \beta_t \eta_t.
\end{equation}
Therefore, we can simplify \eqref{eqn:proof-beta-update} to
\begin{equation}\label{eqn:proof-beta-update-simplified}
\begin{multlined}
    L_{\beta_t}(\mtx{W}_{t+1},\vct{y}_t) - \mathrm{Tr}(\mtx{C}\mtx{W}_\star)
    \leq (1-\eta_t) \Big( L_{\beta_{t-1}}(\mtx{W}_t,\vct{y}_t) - \mathrm{Tr}(\mtx{C}\mtx{W}_\star) \Big) \\
    + \frac{1}{2} \beta_t \eta_t^2 \norm{\mathcal{A}}^2 p^2. 
\end{multlined}
\end{equation}

Then, we change the dual variable on the left hand side of the inequality to obtain a recursion:
\begin{equation}\label{eqn:proof-dual-variable}
\begin{multlined}
    L_{\beta_t}(\mtx{W}_{t+1},\vct{y}_{t+1}) - \mathrm{Tr}(\mtx{C}\mtx{W}_\star) \\
    ~~~~~~\begin{aligned}
    & = L_{\beta_t}(\mtx{W}_{t+1},\vct{y}_t) - \mathrm{Tr}(\mtx{C}\mtx{W}_\star) + (\vct{y}_{t+1} - \vct{y}_t)^\top (\mathcal{A}\mtx{W}_{t+1} - \vct{v}) \\
    & = L_{\beta_t}(\mtx{W}_{t+1},\vct{y}_t) - \mathrm{Tr}(\mtx{C}\mtx{W}_\star) + \gamma_t \norm{\mathcal{A}\mtx{W}_{t+1} - \vct{v}}^2\\
    & \leq L_{\beta_t}(\mtx{W}_{t+1},\vct{y}_t) - \mathrm{Tr}(\mtx{C}\mtx{W}_\star) + \beta_t \eta_t^2 \norm{\mathcal{A}}^2 p^2,%
    \end{aligned}
\end{multlined}
\end{equation}
where the last line is ensured by the assumptions on the choice of $\gamma_t$. We combine \eqref{eqn:proof-beta-update-simplified} and \eqref{eqn:proof-dual-variable},
\begin{equation}
\begin{multlined}
    L_{\beta_t}(\mtx{W}_{t+1},\vct{y}_{t+1}) - \mathrm{Tr}(\mtx{C}\mtx{W}_\star) 
    \leq (1-\eta_t) \Big( L_{\beta_{t-1}}(\mtx{W}_t,\vct{y}_t) - \mathrm{Tr}(\mtx{C}\mtx{W}_\star) \Big) \\
    + \frac{3}{2} \beta_t \eta_t^2 \norm{\mathcal{A}}^2 p^2. 
\end{multlined}
\end{equation}
Using this recursion for iterations $1.\ldots,t$, we obtain
\begin{equation}
\begin{multlined}
    L_{\beta_t}(\mtx{W}_{t+1},\vct{y}_{t+1}) - \mathrm{Tr}(\mtx{C}\mtx{W}_\star) 
    \leq (1-\eta_1) \Big( L_{\beta_0}(\mtx{W}_1,\vct{y}_1) - \mathrm{Tr}(\mtx{C}\mtx{W}_\star) \Big) \\
    + \frac{3}{2} \norm{\mathcal{A}}^2 p^2 \sum_{i = 1}^t \beta_i \eta_i^2 \prod_{j = i+1}^t (1-\eta_j)  . 
\end{multlined}
\end{equation}
The first term on the right side is 0 since $\eta_1 = 1$. We focus on the second term:
\begin{equation}
\begin{multlined}
     \sum_{i = 1}^t \beta_i \eta_i^2 \prod_{j = i+1}^t (1-\eta_j)
     =  4 \beta_0 \sum_{i = 1}^t \frac{1}{(i+1)^{3/2}} \prod_{j = i+1}^t \frac{j-1}{j+1} 
     \\ = 4 \beta_0 \sum_{i = 1}^t \frac{1}{(i+1)^{3/2}} \frac{i(i+1)}{t(t+1)} 
     \leq \frac{4 \beta_0}{t(t+1)} \sum_{i = 1}^t i^{1/2} 
     \leq \frac{4 \beta_0}{\sqrt{t+1}}.
\end{multlined}
\end{equation}
Hence, we conclude that
\begin{equation}\label{eqn:proof-lagrangian-convergence}
    L_{\beta_t}(\mtx{W}_{t+1},\vct{y}_{t+1}) - \mathrm{Tr}(\mtx{C}\mtx{W}_\star) 
    \leq \frac{6\beta_0 \norm{\mathcal{A}}^2 p^2}{\sqrt{t+1}}. 
\end{equation}

The bound on the objective residual follows immediately from \eqref{eqn:proof-lagrangian-convergence}, since
\begin{equation}\label{eqn:proof-lagrangian-objective}
\begin{aligned}
    L_{\beta_t}(\mtx{W}_{t+1},&\vct{y}_{t+1})
    = \mathrm{Tr}(\mtx{C}\mtx{W}_{t+1}) + \vct{y}_{t+1}^\top (\mathcal{A}\mtx{W}_{t+1} - \vct{v}) + \frac{\beta_t}{2} \norm{\mathcal{A}\mtx{W}_{t+1} + \vct{v}}^2 \\
    & ~~ = \mathrm{Tr}(\mtx{C}\mtx{W}_{t+1}) -  \frac{1}{2\beta_t}\norm{\vct{y}_{t+1}}^2 + \frac{\beta_t}{2}\norm{\mathcal{A}\mtx{W}_{t+1} + \vct{v} - \beta_t^{-1} \vct{y}}^2 \\
    & ~~ \geq \mathrm{Tr}(\mtx{C}\mtx{W}_{t+1}) -  \frac{D^2}{2\beta_t},
\end{aligned}
\end{equation}
where the last line depends on the boundedness assumption on $\vct{y}_t$. We combine \eqref{eqn:proof-lagrangian-convergence} and \eqref{eqn:proof-lagrangian-objective} and get
\begin{equation}\label{eqn:proof-lagrangian-rate}
    \mathrm{Tr}(\mtx{C}\mtx{W}_{t+1}) - \mathrm{Tr}(\mtx{C}\mtx{W}_\star) 
    \leq \frac{6\beta_0 \norm{\mathcal{A}}^2 p^2}{\sqrt{t+1}} + \frac{D^2}{2\beta_0 \sqrt{t+1}}.
\end{equation}

It remains to prove the bound on infeasibility. We revisit \eqref{eqn:proof-lagrangian-rate}, invoke Cauchy-Schwarz inequality and the boundedness assumption on $\vct{y}$ to obtain
\begin{equation}\label{eqn:proof-revisit}
\begin{multlined}
    L_{\beta_t}(\mtx{W}_{t+1},\vct{y}_{t+1}) - \mathrm{Tr}(\mtx{C}\mtx{W}_\star) \\
    = \mathrm{Tr}(\mtx{C}\mtx{W}_{t+1}) - \mathrm{Tr}(\mtx{C}\mtx{W}_\star) + \vct{y}_{t+1}^\top (\mathcal{A}\mtx{W}_{t+1} - \vct{v}) + \frac{\beta_t}{2} \norm{\mathcal{A}\mtx{W}_{t+1} + \vct{v}}^2 \\
    \leq \frac{6\beta_0 \norm{\mathcal{A}}^2 p^2}{\sqrt{t+1}}.
\end{multlined}
\end{equation}

Based on the strong duality assumption, we use the Lagrangian saddle point theory \cite[Section~5.4]{boyd2004convex}, 
\begin{equation}\label{eqn:proof-lagrange-duality}
\begin{multlined}
    \underbrace{\mathrm{Tr}(\mtx{C}\mtx{W}_\star)}_{\displaystyle L_0(\mtx{W}_\star,\vct{y}_\star)} \leq \underbrace{\mathrm{Tr}(\mtx{C}\mtx{W}_{t+1}) + \vct{y}_\star^\top (\mathcal{A}\mtx{W}_{t+1} - \vct{v})}_{\displaystyle L_0(\mtx{W}_{t+1},\vct{y}_\star)}. 
\end{multlined}
\end{equation}
We combine \eqref{eqn:proof-revisit} and \eqref{eqn:proof-lagrange-duality}:
\begin{equation}
\begin{multlined}
    (\vct{y}_{t+1} - \vct{y}_\star)^\top (\mathcal{A}\mtx{W}_{t+1} - \vct{v}) + \frac{\beta_t}{2} \norm{\mathcal{A}\mtx{W}_{t+1} + \vct{v}}^2 \leq \frac{6\beta_0 \norm{\mathcal{A}}^2 p^2}{\sqrt{t+1}}.
\end{multlined}
\end{equation}
We use Cauchy-Schwarz and the boundedness assumption on $\vct{y}$ to obtain a second-order inequality of $\norm{\mathcal{A}\mtx{W}_{t+1} - \vct{v}}$:
\begin{equation}
\begin{multlined}
    -2D \norm{\mathcal{A}\mtx{W}_{t+1} - \vct{v}} + \frac{\beta_t}{2} \norm{\mathcal{A}\mtx{W}_{t+1} + \vct{v}}^2 \leq \frac{6\beta_0 \norm{\mathcal{A}}^2 p^2}{\sqrt{t+1}}.
\end{multlined}
\end{equation}
By solving this inequality for $\norm{\mathcal{A}\mtx{W}_{t+1} - \vct{v}} \geq 0$, we get
\begin{equation}
    \norm{\mathcal{A}\mtx{W}_{t+1} - \vct{v}} \leq \frac{1}{\beta_t} \left( 4D + 2\sqrt{3} \beta_0 p  \norm{\mathcal{A}} \right).
\end{equation}

\subsection{On the Dual Step-size of FWAL}
Theoretical analysis of FWAL depends on the assumption that the dual step-size $\gamma_t \geq 0$ satisfies
\begin{equation}\label{eqn:gamma-condition-1}
    \gamma_t \norm{\vct{g}_t}^2 \leq \beta_t \eta_t^2 p^2 \norm{\mathcal{A}}^2
\end{equation}
and the bounded travel condition $\norm{\vct{y}_{t+1}} \leq D$. Note, the largest step-size that satisfies this condition can be computed analytically. First, we take the square,
\begin{equation}
\begin{multlined}
    \norm{\vct{y}_{t+1}}^2 = \norm{\vct{y}_t + \gamma_t \vct{g}_t}^2 = \norm{\vct{y}_t}^2 + 2\gamma_t \vct{y}_t^\top \vct{g}_t + \gamma_t^2 \norm{\vct{g}_t}^2 \leq D^2.
\end{multlined}
\end{equation}
Then, by solving this inequality for $\gamma_t \geq 0$, we obtain
\begin{equation}\label{eqn:gamma-condition-2}
    \gamma_t \leq \frac{- \vct{y}_t^\top \vct{g}_t + \sqrt{(\vct{y}_t^\top \vct{g}_t)^2 + (D^2 - \norm{\vct{y}_t}^2)\norm{\vct{g}_t}^2}}{\norm{\vct{g}_t}^2}.
\end{equation}
Combining \eqref{eqn:gamma-condition-1} and \eqref{eqn:gamma-condition-2}, we can choose 
\begin{equation}
    \gamma_t \leq \min \left\{\frac{\beta_t \eta_t^2 p^2 \norm{\mathcal{A}}^2}{\norm{\vct{g}_t}^2},  \frac{- \vct{y}_t^\top \vct{g}_t + \sqrt{(\vct{y}_t^\top \vct{g}_t)^2 + (D^2 - \norm{\vct{y}_t}^2)\norm{\vct{g}_t}^2}}{\norm{\vct{g}_t}^2}\right\}
\end{equation}
and $\gamma_t = 0$ if $\norm{\vct{g}_t} = 0$. Prior work \cite{yurtsever2019conditional,yurtsever2021scalable} invoke also the fixed threshold $\gamma_t \leq \beta_0$ to avoid very large steps when $\norm{\vct{g}_t}$ is small:
\begin{equation}\label{eqn:gamma-conditions}
    \gamma_t \leq \min \left\{\beta_0, \frac{\beta_t \eta_t^2 p^2 \norm{\mathcal{A}}^2}{\norm{\vct{g}_t}^2},  \frac{- \vct{y}_t^\top \vct{g}_t + \sqrt{(\vct{y}_t^\top \vct{g}_t)^2 + (D^2 - \norm{\vct{y}_t}^2)\norm{\vct{g}_t}^2}}{\norm{\vct{g}_t}^2}\right\}.
\end{equation}
Note, $\gamma_t = 0$ always satisfies this condition hence is a valid choice. In fact, FWQP is a special case of FWAL with $\vct{y}_0 = \vct{0}$ and $\gamma_t = 0$. 

In numerical experiments, we use a constant step-size $\gamma_t = \beta_0$. This choice may fail the conditions in  \eqref{eqn:gamma-conditions} but works well in practice. 

\subsection{Inequality Constraints}
The problems addressed in our paper are concerned with equality constraints. However, many problems such as resolving \emph{partial} permutations might require us to naturally handle inequalities. In this section we present one possible way to accommodate affine inequality constraints in Q-FW and leave it as a future study to experiment on tasks with such constraints. Our particular solution requires the evaluation of D-Wave Quantum Computer only as many times as in the case of equality constraints. 

Without loss of generality, we assume that the inequality constraints are given in the form of
\begin{equation}\label{eqn:inequality_constraint_upper}
    \vct{e}_i^\top \vct{x} \leq f_i , \quad i = 1, 2, \ldots, q.
\end{equation}
Since $\vct{x}$ is binary valued, we can derive trivial lower and upper bounds
\begin{equation}\label{eqn:inequality_constraint_lower}
\begin{gathered}
    -\alpha_i \leq \vct{e}_i^\top \vct{x} \leq \beta_i, ~~\text{where}~~  \\
    \alpha_i = -\sum_{j=1}^n \min\{(\vct{e}_i)_j, 0\}, ~~\text{and}~~ \beta_i = \sum_{j=1}^n \max\{(\vct{e}_i)_j, 0\}.
\end{gathered}
\end{equation}
In other words, $\alpha_i$ is the sum of absolute values of the negative coefficients of $\vct{e}_i$, and $\beta_i$ is the sum of its positive coefficients. By definition, $\alpha_i$ and $\beta_i$ are nonnegative. We assume $f_i < \beta_i$, because otherwise the constraint is redundant and we can remove it. We also assume that $-\alpha_i \leq f_i$. Otherwise, the feasible set is empty and there are no solutions. 

We combine \eqref{eqn:inequality_constraint_upper} and \eqref{eqn:inequality_constraint_lower}, add $\alpha_i$ to both sides:
\begin{equation}\label{eqn:linear-inequality-terms}
    0 \leq \vct{e}_i^\top \vct{x} + \alpha_i \leq f_i + \alpha_i, \quad i = 1, 2, \ldots, q.
\end{equation}
Since all sides are nonnegative, we can now take the squares and get
\begin{equation}
    0 \leq (\vct{e}_i^\top \vct{x})^2 + \alpha_i^2 + 2\alpha_i (\vct{e}_i^\top \vct{x}) \leq (f_i + \alpha_i)^2, \quad i = 1, 2, \ldots, q. 
\end{equation}
Then, we replace $(\vct{e}_i^\top \vct{x})^2 = \mathrm{Tr}(\vct{x}^\top \vct{e}_i \vct{e}_i^\top \vct{x}) = \mathrm{Tr}(\vct{e}_i \vct{e}_i^\top\vct{x} \vct{x}^\top )$ by $\mathrm{Tr}(\mtx{E}_i \mtx{X})$ where $\mtx{E}_i = \vct{e}_i\vct{e}_i^\top$. We subtract $\alpha_i^2$ and get
\begin{equation} \label{eqn:quadratic-inequality-terms}
    -\alpha_i^2 \leq \mathrm{Tr}(\mtx{E}_i \mtx{X}) + 2\alpha_i(\vct{e}_i^\top \vct{x}) \leq f_i^2 + 2\alpha_i f_i, \quad i = 1, 2, \ldots, q.
\end{equation}

By combining \eqref{eqn:linear-inequality-terms} and \eqref{eqn:quadratic-inequality-terms}, we reformulate $q$ inequality constraints of the original QBO problem as $2q$ inequality (box) constraints in our CP relaxation:
\begin{equation}
\begin{aligned}
    & -\alpha_i \leq \vct{e}_i^\top \vct{x} \leq f_i, & & \quad i = 1,2,\ldots, q\\
    & -\alpha_i^2 \leq \mathrm{Tr}(\mtx{E}_i \mtx{X}) + 2\alpha_i(\vct{e}_i^\top \vct{x}) \leq f_i^2 + 2 \alpha_i f_i, & & \quad i = 1, 2, \ldots, q
\end{aligned}
\end{equation}

Finally, we introduce a linear map $\mathcal{E}:\mathbb{R}^{p\times p} \to \mathbb{R}^{2q}$ and two vectors $\boldsymbol{l},\vct{u} \in \mathbb{R}^{2q}$ to simplify the notation to $\boldsymbol{l} \leq \mathcal{E}(\mtx{W}) \leq \vct{u}$, or explicitly,
\begin{equation}\label{eqn:Eoperator}
    \underbrace{\begin{bmatrix} 
    -\alpha_1 \\
    \vdots \\
    -\alpha_q \\
    -\alpha_1^2 \\
    \vdots \\
    -\alpha_q^2
    \end{bmatrix}}_{\displaystyle\boldsymbol{l}} 
    \leq 
    \underbrace{\begin{bmatrix} 
    \vct{e}_1^\top \vct{x} \\
    \vdots \\
    \vct{e}_q^\top \vct{x} \\
    \mathrm{Tr}(\mtx{E}_1\vct{X}) + 2 \alpha_1 (\vct{e}_1^\top \vct{x}) \\
    \vdots \\
    \mathrm{Tr}(\mtx{E}_q\vct{X}) + 2 \alpha_q (\vct{e}_q^\top \vct{x})
    \end{bmatrix}}_{\displaystyle\mathcal{E}(\mtx{W})} 
    \leq 
    \underbrace{\begin{bmatrix} 
    f_1 \\
    \vdots \\
    f_q \\
    f_1^2 + 2\alpha_1 f_1 \\
    \vdots \\
    f_q^2 + 2\alpha_q f_q
    \end{bmatrix}}_{\displaystyle\vct{u}} 
\end{equation}
where the inequalities are entrywise.

Next, we present FWAL steps for inequality constraints. This extension is detailed in \cite[Section~D.4]{yurtsever2021scalable}, we present it here for completeness. We use the following augmented Lagrangian formulation to derive FWAL steps:
\begin{equation}
\begin{multlined}
    L_{\beta}(\mtx{W};\vct{y};\vct{y}') = \mathrm{Tr}(\mtx{C}\mtx{W}) + \vct{y}^\top (\mathcal{A}\mtx{W} - \vct{v}) + \frac{\beta}{2} \norm{\mathcal{A}\mtx{W} - \vct{v}}^2 \\ + \min_{\boldsymbol{l} \leq \boldsymbol{\omega} \leq \vct{u}} \left\{ \vct{y}'^\top (\mathcal{E}\mtx{W} - \boldsymbol{\omega}) + \frac{\beta}{2} \norm{\mathcal{E}\mtx{W} - \boldsymbol{\omega}}^2 \right\}.
\end{multlined}
\end{equation}
Then, the partial derivative of $L_{\beta_t}$ with respect to the first variable is 
\begin{equation}
\begin{multlined}
    \mtx{G}_t = \mtx{C} + \mathcal{A}^\top \vct{y}_t + \beta_t \mathcal{A}^\top (\mathcal{A}\mtx{W}_t - \vct{v}) + \mathcal{E}^\top \vct{y}'_t + \beta_t \mathcal{E}^\top (\mathcal{E}\mtx{W}_t - \boldsymbol{\omega}^\star_t), \\
    \text{where}~ \boldsymbol{\omega}^\star_t  = \arg\min_{\boldsymbol{l} \leq \boldsymbol{\omega} \leq \vct{u}} \left\{ {\vct{y}_t'}^\top (\mathcal{E}\mtx{W}_t - \boldsymbol{\omega}) + \frac{\beta_t}{2} \norm{\mathcal{E}\mtx{W}_t - \boldsymbol{\omega}}^2 \right\}.
\end{multlined}
\end{equation}
The $\boldsymbol{\omega}^\star_t$ subproblem amounts to a projection, which in turn is a clipping (thresholding) operator:
\begin{equation}
    \begin{aligned}
    \boldsymbol{\omega}^\star_t & = \arg\min_{\boldsymbol{l} \leq \boldsymbol{\omega} \leq \vct{u}} \left\{ {\vct{y}_t'}^\top (\mathcal{E}\mtx{W}_t - \boldsymbol{\omega}) + \frac{\beta_t}{2} \norm{\mathcal{E}\mtx{W}_t - \boldsymbol{\omega}}^2 \right\} \\
    & = \arg\min_{\boldsymbol{l} \leq \boldsymbol{\omega} \leq \vct{u}} \left\{ \norm{\mathcal{E}\mtx{W}_t - \boldsymbol{\omega} + \beta_t^{-1}\vct{y}_t'}^2 \right\} \\ 
    & = \mathrm{proj}_{[\boldsymbol{l},\vct{u}]} (\mathcal{E}\mtx{W}_t + \beta_t^{-1}\vct{y}_t') := \mathrm{clip} (\mathcal{E}\mtx{W}_t + \beta_t^{-1}\vct{y}_t', \boldsymbol{l}, \vct{u}).
\end{aligned}
\end{equation}

The update rule for the dual variable $\vct{y}$ remains the same. Similarly, for $\vct{y}'$, we take a small gradient ascent step by using the partial derivative of $L_\beta$ with respect to the third variable, 
\begin{equation}
    \vct{g}_t' = \mathcal{E} \mtx{W}_{t+1} - \boldsymbol{\omega}^\star_t, \quad \text{and} \quad \vct{y}_{t+1}' = \vct{y}_t' + \gamma_t \vct{g}_t'.
\end{equation}

\subsection{Early Stopping Heuristics}
As D-Wave provides a limited amount of computation, we are bound to use our resources wisely. To this end, for some of the synchronization experiments\footnote{usually for QPS, we solve larger problems than QGGM}, we opt for (i) a faster update, (ii) an automatic termination when good quality solutions are found. 
We take a different approach and propose two modifications to the original Q-FWAL:
\begin{enumerate}[noitemsep,leftmargin=\parindent,topsep=1pt]
    \item projecting the solution to the feasibility set at each iteration and switching the current solution with the projected, if:
    \begin{equation}
    \mathrm{Tr}(\mtx{C}\hat{\mtx{H}}_t) < \mathrm{Tr}(\mtx{C}\W_t)
    \end{equation}
    where $\hat{\mtx{H}}_t$ is obtained by lifting the rounded, intermediate solution at time $t$, \ie for permutations, applying Hungarian algorithm on the left singular vectors of $\X_t$. 
    \item the stopping criteria that checks the constraints are satisfied and the cost remains unchanged in consecutive iterations:
    \begin{align}
        \|\mathcal{A}\W_t-\vct{v}\|=0 \quad \mathrm{\, and\, }\quad \mathrm{Tr}(\mtx{C}\W_t)=\mathrm{Tr}(\mtx{C}\W_{t-1})
    \end{align}
\end{enumerate}
Note that, typical Frank Wolfe-type algorithms usually make use of the \emph{duality gap} as a practical \emph{stopping criterion} motivated by the fact that this quantity upper bounds the primal gap while at the same time enjoying the same asymptotic guarantees.~\cite{frandi2015partan,jaggi2013revisiting,yurtsever2021scalable}. However, we find that in practice this is still a very soft barrier, satisfied only at high number of iterations. This is the reason why we preferred the two proposed modifications above.

\subsection{Psuedocode}
We are now ready to provide the pseudocode for Q-FW. In the main paper we always use the equality constraints as these are the most common in the tasks we address. However, for the sake of generality we present in Alg.~\ref{alg:QFW} the generic Q-FW approach for handling inequality and equality constraints. We will make our implementation available upon publication.
\begin{algorithm}
\setstretch{1.13}
\caption{Q-FW for Quadratic Binary Optimization.}\label{alg:QFW}
 \textbf{Input}: {Cost matrix $\mtx{Q} \in \mathbb{R}^{n \times n}$,\;
 Equality constraints $\{(\vct{a}_i,b_i)\}_{i=1}^m$,\; 
 Inequality constraints $\{(\vct{e}_i,f_i)\}_{i=1}^q$},\;\# of iterations $T$,\;penalty parameter $\beta_0 > 0$ (default $1$)\\[0.25em]
 \textbf{Preparation}: {$p \gets n+1,~d\gets 2m+n+1,~d'\gets 2q$. 
 Form $\smash{\mtx{C} \gets \begin{brsm} 0 & \vct{0}^\top \\ \vct{0} & \vct{Q} \end{brsm}}$.
 Construct $(\mathcal{A}, \vct{v})$ as defined in \eqref{eqn:Aoperator}, and $(\mathcal{E}, \boldsymbol{l},\vct{u})$ as in \eqref{eqn:Eoperator}.}\\[0.5em]
 \textbf{Initialization}: 
 $\mtx{W} \gets \mtx{0}^{p \times p},$\;
 $\vct{y} \gets \vct{0}^d,$\;
 $\vct{y}' \gets \vct{0}^{d'}$\;\\[0.5em]
\textbf{Main loop} [FWAL]:\\
\For{$t = 1,\ldots,T$}{
$\eta \gets 2/(t+1),$~~and~~$\beta \gets \beta_0 \sqrt{t+1}$\\
$\vct{g} \gets \mathcal{A}\mtx{W} - \vct{v},$~~and~~$\vct{g}' \gets \mathcal{E}\mtx{W} - \mathrm{clip}(\mathcal{E}\mtx{W}+\beta^{-1}\vct{y}',\boldsymbol{l},\vct{u})$\\
$\mtx{G} \gets \mtx{C} + \mathcal{A}^\top (\vct{y} +  \beta \vct{g}) + \mathcal{E}^\top (\vct{y}' + \beta \vct{g}')$\\
$\vct{w} \gets \arg\min \; \{ \vct{w}^\top \mtx{G} \vct{w} : ~ \vct{w} \in \mathbb{Z}_2^p \}$ \Comment{QUBO subproblem}\\
$\mtx{W} \gets (1-\eta)\mtx{W} + \eta \vct{ww}^\top$ \\
$\vct{g} \gets \mathcal{A}\mtx{W} - \vct{v},$~~and~~$\vct{g}' \gets \mathcal{E}\mtx{W} - \mathrm{clip}(\mathcal{E}\mtx{W}+\beta_+^{-1}\vct{y}',\boldsymbol{l},\vct{u})$\Comment{$\beta_+ = \beta_0 \sqrt{t+2}$}\\
$\vct{y} \gets \vct{y} + \gamma \vct{g},$~~and~~$\vct{y}' \gets \vct{y}' + \gamma \vct{g}'$~~\Comment{In practice, we use $\gamma = \beta_0$}
}

\vspace{1em}

\textbf{Main loop} [FWQP]:\\
\For{$t = 1,\ldots,T$}{
$\eta \gets 2/(t+1),$~~and~~$\beta \gets \beta_0 \sqrt{t+1}$\\
$\vct{g} \gets \mathcal{A}\mtx{W} - \vct{v},$~~and~~$\vct{g}' \gets \mathcal{E}\mtx{W} - \mathrm{clip}(\mathcal{E}\mtx{W},\boldsymbol{l},\vct{u})$\\
$\mtx{G} \gets \mtx{C} + \beta \mathcal{A}^\top \vct{g} + \beta \mathcal{E}^\top \vct{g}'$\\
$\vct{w} \gets \arg\min \; \{ \vct{w}^\top \mtx{G} \vct{w} : ~ \vct{w} \in \mathbb{Z}_2^p \}$ \Comment{QUBO subproblem}\\
$\mtx{W} \gets (1-\eta)\mtx{W} + \eta \vct{ww}^\top$ 
}

\vspace{0.5em}
\textbf{Rounding}: (Option 1) Extract $\vct{x}$ by taking the first column of $\mtx{W}$ and removing its first entry. 
(Option 2) Extract $\mtx{X}$ by removing the first row and first column of $\mtx{W}$. Compute $\vct{x}$ as the top singular vector of $\mtx{X}$.
-- Project $\vct{x}$ onto $\mathbb{Z}_2^n$.\\[0.5em]
\textbf{Output}: Solution $\mtx{W} \in \Delta^p$ for the copositive program, and $\vct{x} \in \mathbb{Z}_2^n$ for the QBO.
\end{algorithm}\vspace{-2mm}
\vspace{-3mm}\section{Adiabatic Quantum Computing}\label{sec:suppAQC}\vspace{-3mm}

\textit{Adiabatic quantum computing} (AQC) is only one of the two quantum computing models. 
AQC and gate-based quantum computing paradigms are said to be polynomially equivalent, in theory (experimental confirmations are ongoing). 
In the gate-based model, all computations on qubits can be represented as unitary  transformations (that can potentially cover the entire Hilbert space); hence, all operations before qubit measurements are invertible. 
AQC model, instead, is defined in terms of Hamilton operator evolution. 
Note, QA can be performed both in an adiabatic and non-adiabatic manner (faster than what the adiabatic theorem dictates).
Current AQC implementations such as DWave \cite{Dattani2019} implement QA, and the quantum system evolution is not guaranteed to be adiabatic. For a more comprehensive overview of the AQC foundations, see \cite{KadowakiNishimori1998,Farhi2001,DWAVE}. 

\begin{figure}[t!]
     \centering
         \centering
         \includegraphics[width=0.9\textwidth]{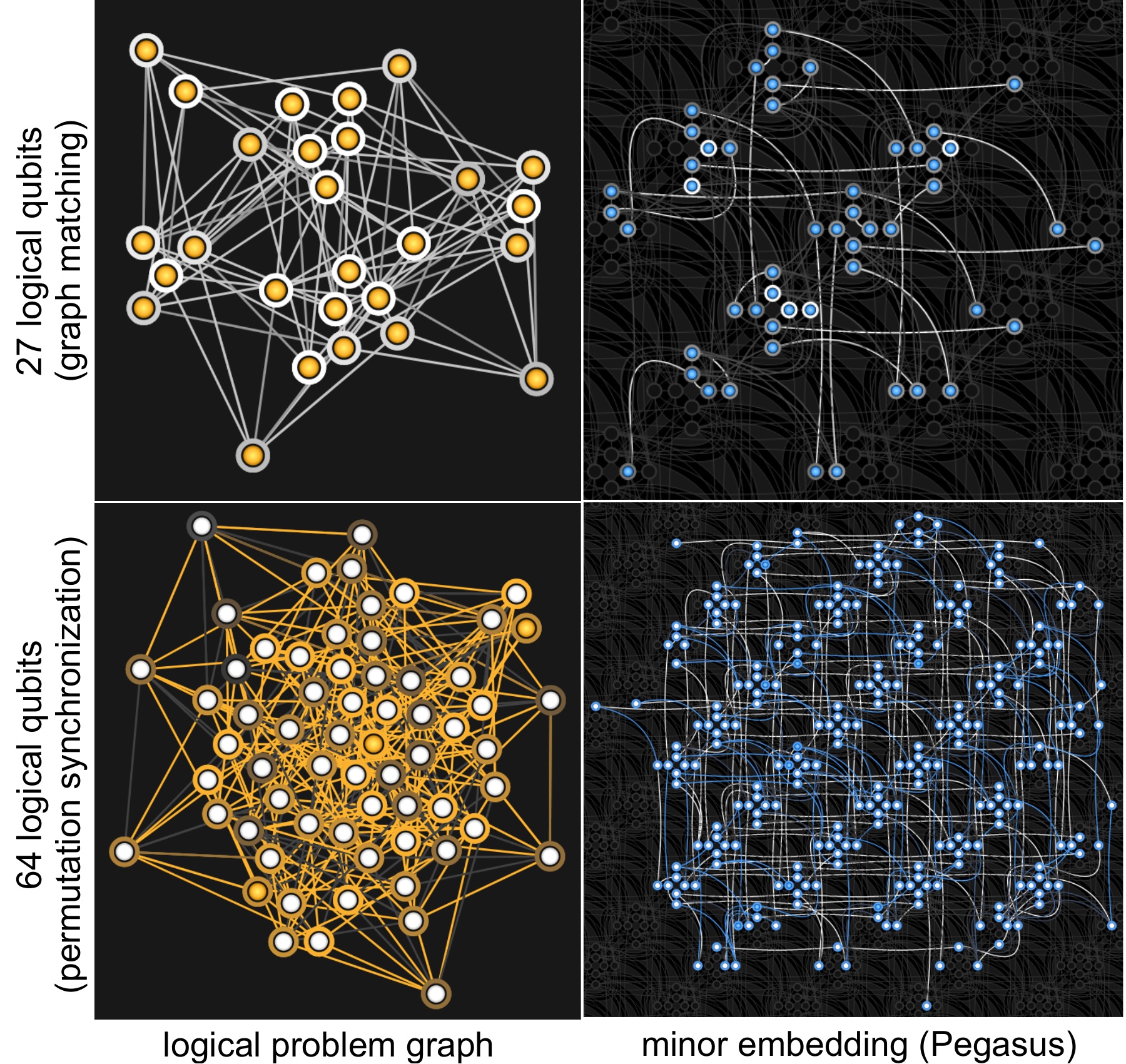}
         \caption{Graphs of the logical problems (the left column) arising in our experiments with $27$ (the top row) and $64$ logical qubits (the bottom row), along with their minor embeddings on the Pegasus topology \cite{Dattani2019} obtained by Cai \textit{et al.}'s method \cite{Cai2014} (the right column). Each node in the logical problem graph represents a logical qubit, and each edge stands for couplings between the logical qubits. Physical qubits build chains in the minor embedding to represent a single logical qubit. 
         } 
         \label{fig:embed}\vspace{-4mm}
\end{figure}

The weight matrix of a QUBO problem defines a \textit{logical} problem \cite{benkner2021q}, \textit{i.e.,} each its binary variable is said to be a \textit{logical} qubit in the idealised quantum hardware context. 
Every logical problem is abstracted from real quantum hardware and  assumes arbitrary connectivity patterns between the qubits. 
This contrasts with the notion of \textit{physical} qubits,  \textit{i.e.,} qubits available in hardware with their connectivity  patterns. 
Since physical qubits are not arbitrarily connected to each other on modern AQCs, multiple of them are required to represent a single logical problem  qubit \cite{Dattani2019}. 
Finding a mapping of a logical QUBO problem to the hardware qubit graph is known as  \emph{minor embedding}; it can be performed with such algorithms as Cai \textit{et al.} \cite{Cai2014}. We give an example involving logical and embedded graphs of two of our problems in~\cref{fig:embed}.
We now briefly describe quantum annealing.

A QUBO optimization is equivalent to minimizing the energy of a classical Ising Hamiltonian $\J$ with no bias field, where the variables $\s_i$ are interpreted as classical spin values. Hence, the minimum of the QUBO objective is equivalently obtained as the \emph{ground state} of Quantum Ising Hamiltonian:
\begin{align}
    \mathcal{H}_P = \sum\limits_{ij} J_{ij} \sigma_P^{(i)}\sigma_P^{(j)}
, \end{align}
where $\sigma_P^{(i)}$ denotes the \emph{Pauli matrix} applied to the $i^\mathrm{th}$ qubit of an $n$-qubit system.
In contrast to a classical bit, a qubit $\ket{\psi}$ can continuously transition between the states $\ket{0}$ and $\ket{1}$ (the equivalents of classical states $0$ and $1$) fulfilling the equation $\ket{\psi} = \alpha \ket{0} + \beta \ket{1}$, with probability amplitudes satisfying $|\alpha|^2 + |\beta|^2 = 1$. 
The eigenvalues of the Hamiltonian correspond to the possible  system's energies. 
As such, the same minimization can be written as:
\begin{align}\label{eq:QH}
    \min_{\ket{\psi}\in\mathbb{C}^{2^n}}\bra{\psi} \mathcal{H}_P \ket{\psi}. 
\end{align}
Adiabatic Quantum Annealing (AQA) solves~\cref{eq:QH} by evolving the Hamiltonian to one where the ground state corresponds to the optimal solution:
\begin{equation}
    \mathcal{H}(\tau) = [1 - \tau]\,\mathcal{H}_I + \tau\,\mathcal{H}_P, 
\end{equation}  
with $\mathcal{H}_I$ being an \emph{initial} Hamiltonian realized as a superposition with equal probabilities of measuring $\ket{0}$ or $\ket{1}$ for every qubit. 
The adiabatic theorem of quantum mechanics \cite{BornFock1928} implies that if a system  transits \textit{gradually enough} (the concrete meaning of  \textit{gradually} depends on many factors), then the system will continue to stay in its ground state in the course of the entire evolution. 
Hence, by the end of the transition, the system will be measured in the ground state of the problem Hamiltonian, \textit{i.e.,} the global optimiser. 

A hybrid algorithm involving QA always has multiple steps that  cover the preparation of a QUBO problem, minor embedding, 
a series of anneals, problem unembedding (from the graph of physical qubits to the logical problem graph), solution selection and solution interpretation. 

\end{appendices}

\end{document}